\documentclass[10pt,twocolumn]{article} 
\usepackage{simpleConference}
\usepackage{times}
\usepackage{graphicx}
\usepackage{amssymb}
\usepackage{url,hyperref}
\usepackage{microtype}

\usepackage{subcaption}

\usepackage{color,soul}

\usepackage{algorithm}
\usepackage{algpseudocode}
\usepackage{booktabs}

\usepackage{blindtext}
\usepackage{siunitx}
\usepackage{amsmath}
\usepackage{authblk}
\usepackage{siunitx}
\usepackage{amsmath}
\usepackage{amssymb}
\usepackage[section]{placeins}
\usepackage{mathtools}
\usepackage{amsthm}
\usepackage{multicol}
\usepackage{dcolumn}
\usepackage{enumitem}
\usepackage{authblk}
\usepackage[capitalize,noabbrev]{cleveref}
\theoremstyle{plain}

\theoremstyle{definition}

\theoremstyle{remark}

\usepackage[textsize=tiny]{todonotes}

\begin{document}

\title{Trend analysis and forecasting air pollution in Rwanda}
\author[1,2]{Paterne Gahungu}
\author[1]{Jean Remy Kubwimana}
\affil[1]{African Institute for Mathematical Sciences, Rwanda}
\affil[2]{Institute of Applied Statistics, University of Burundi}

\maketitle
\thispagestyle{empty}

\vskip 0.3in

\setlist[itemize]{noitemsep, topsep=0pt}

\setlist[enumerate]{noitemsep, topsep=0pt}

\begin{abstract}

Air pollution is a major public health problem worldwide although the lack of data  is a global issue for most low and middle income countries. Ambient air pollution in the form of fine particulate matter (PM2.5) exceeds the World Health Organization guidelines in Rwanda with a daily average of around $\SI{42.6}{\micro\gram}/m^3$.  Monitoring and mitigation strategies require an expensive investment in equipment to collect pollution data. Low-cost sensor technology and machine learning methods have appeared as an alternative solution to get reliable information for decision making. This paper analyzes the trend of air pollution in Rwanda and proposes forecasting models suitable to data collected by a network of low-cost sensors deployed in Rwanda. 

\end{abstract}

\section{Introduction }

Pollution of ambient air in the form of particulate matter (PM2.5) is associated with negative impact on human health \cite{kim2015review}\cite{wong2015satellite}\cite{rajak2020short}\cite{shehab2019effects}. Diseases like asthma and lung cancer are worsened by exposure to air pollution.  According to the World Health Organization (WHO), air pollution is responsible for about seven million premature deaths worldwide per year \cite{weltgesundheitsorganisation2021global}. The African continent is one of the most affected region with high levels of air pollution  and its health damages with an estimated number of 780 000 premature deaths \cite{bauer2019desert}. In Rwanda, the estimates are about  three thousand deaths in 2016 \cite{brauer2012exposure}. These estimates are very uncertain as the inference is done using data from satellite observations. The lack of ground-based data on air pollution is primarily due to the high cost of equipment used in monitoring. The reference monitors like the  Met One Beta Attenuation Mass Monitor cost hundreds of thousands of dollars which is not affordable for most African countries. Effective policies around air pollution control are highly dependent on the availability of data. Countries are trying to formulate and implement measures to reduce air pollution: the car-free Sundays in Rwanda and the increase in duties for older imported cars in the East African economic region since 2017. However, to track  the effect of pollution-control strategies, countries need data for a long period of time.

Low-cost sensors are being utilized to fill in this gap in air pollution data in low and middle-income countries. However, low-cost sensors are very sensitive to weather conditions such as temperature and relative humidity. Low-cost sensors, when calibrated, can provide accurate data that can be used for policy planning.

There are few published works on air pollution in Rwanda \cite{kalisa2018characterization}\cite{subramanian2020air}. In \cite{kalisa2018characterization}, data was collected on PM2.5 and PM10 in a short-term campaign (a three-month period) for both urban background and rural areas. The study in \cite{kalisa2018characterization} used standard filter-based gravimetry. Samples were collected daily at three locations in Rwanda: an urban background area away from major pollution sources like industries and high road-traffic, an area near a road to account for transport related pollution  and a rural area far away from Kigali. The work in \cite{subramanian2020air} focuses on spatial and temporal variability of air pollution in Kigali with data collected by a limited number of low-cost sensors. However, to the best of our knowledge, there are no long-term studies on air pollution in Rwanda.

In this work, we present a trend analysis and forecast of air pollution in Rwanda using data obtained from the Rwanda Environmental Management Authority (REMA). Our analysis considers seasonal, monthly, weekly, daily and hourly data. Qualitative information on air possible pollution sources in Rwanda are obtained. Three different methods are used to forecast air pollution in Rwanda: Auto-regressive integrated moving average (ARIMA), Artificial Neural Network (ANN) and Gaussian Processes (GP).
\section{Data and Methods}
\subsection{Data}
The air pollution data used in this study was collected by low-cost sensor monitors  deployed in Kigali since 2018.   Ambient air pollutants measured in Rwanda include PM2.5, SO2, NO2, CO. This study focuses on  particulate matter with diameter less than 2.5 micro-meter (PM2.5). The sensors are manufactured by SENSIT Technologies, LLC (www.gasleaksensors.com). 
\begin{table}[ht!]
	\centering
	\caption{Air Quality Monitoring Stations}
	\begin{tabular}{lccr}
		\textbf{Stations} \\
		
		Gitega        &  Rusororo  \\
		Gacuriro    & Kiyovu \\
		Rebero & Mount Kigali \\
		Kimihurura &  Gikondo Mburabuturo \\
		Gikomero  \\
		
	\end{tabular}
	\label{80}
\end{table}

\subsection{Methods}
We give a brief overview of time series models used in this work. These widely used models include the traditional statistical ARIMA model and current state-of-the-art machine learning algorithms: Neural Network and Gaussian Processes. 
\subsubsection{ARIMA Model}
%
We are interested in predicting $Y_t$ given its historical values $Y_{t-1}, \cdots, Y_0$. An AUto-Regressive (AR) model is given as:
\begin{align}\label{2}
	Y_t= \alpha+ \beta_1 Y_{t-1}+\beta_2 Y_{t-2}+\cdots+ \beta_{p} Y_{t-p}+ \epsilon_t
\end{align}
	\begin{align*}
	Y_t=\alpha + \sum_{i=1}^{p}\beta_iY_{t-p}+\epsilon_t
\end{align*}
where $\beta$=($\beta_{1},\cdots,\beta_p$) is the vector of model coefficients and p is a non-negative integer at any lag, $\alpha$ is the intercept of the model and $\epsilon_{t}$ is the white noise with zero mean and $\sigma^2$ as variance ($\epsilon_{t}\sim \mathcal{N}(0,1)$).

In this model, correlation is introduced between the random variables by regressing $Y_t$ on past values $Y_{t-1},\cdots,Y_{t-p}$.
Moving Average (MA) \cite{abhilash2018time} is a mathematical model in which $Y_t$ is determined solely by the lagged forecast errors. Let $W_t \underset{iid}{\sim}N(0,\sigma_W^2)$ be the white noise which is identically and independently normally distributed with zero mean and the same variance. Then, a moving average of order 1 (MA(1)) and order 2 (MA(2)) are written in the following way: 
\begin{align*}
	Y_t= \mu+W_t+\theta_1W_{t-1}
\end{align*} 
\begin{align*}
	Y_t= \mu+W_t+\theta_1W_{t-1}+\theta_2W_{t-2}
\end{align*} 
Therefore, a moving average model of order $q$ (MA(q)) is defined by the equation written below:
\begin{align}\label{3}
	Y_t=\mu+W_t+\theta_1W_{t-1}+\theta_2W_{t-2}+\cdots+\theta_qW_{t-q}
\end{align} 
\begin{align*}
	W_t=& \sigma \times \epsilon_{t}\\
	\epsilon_{t} \sim N(0,1)
\end{align*}
where $W_t$ shock for the process, and $\sigma$ is the conditional standard deviation. The output ($Y_t$) is determined by long run average $(\mu)$ and weighted sum of past shocks ($W_t$).

The moving average model is stable and has a finite long-run mean ($\mu$) and variance.
\begin{itemize}
	\item[(i)] The unconditional mean
	\begin{align*}
		E(Y_t)=& E(\mu+W_t+\theta_1W_{t-1}+\cdots+\theta_qW_{t-q})\\
		E(Y_t)= &\mu
	\end{align*}
	\item[(ii)]The unconditional variance
	\begin{align*}
		Var(Y_t)=& E\left[(Y_t-\mu) (Y_t-\mu)\right]\\
		Var(Y_t)=&E\left[(W_t+\theta_1W_{t-1}+\cdots+\theta_qW_{t-q})^2\right]
	\end{align*}
	\begin{align*}
		Var(Y_t)= (1+\theta_1^2+\theta_2^2+\cdots+\theta_q)\sigma^2
	\end{align*}
\end{itemize}
Therefore, the AutoRegressive Integrated Moving Average model is found by combining the autoregression (\ref{2}) and moving average (\ref{3}) models.
\begin{multline}
	Y_t= \alpha+ \beta_1 Y_{t-1}+\beta_2 Y_{t-2}+\cdots+ \beta_{p} Y_{t-p}+ \epsilon_t+\mu+W_t+\\\theta_1W_{t-1}+\theta_2W_{t-2}+\cdots+\theta_qW_{t-q}
\end{multline}

\subsubsection{Neural Network} 

We are given observations  $Y_{t-1}, \cdots, Y_0$ and the objective is to predict  $Y_{t}$. 
\begin{eqnarray}
	Y_t=f(\sum_{i=0}^{t-1} w_iY_i+b)
\end{eqnarray}

where $f$ is an activation function,  $w_0, \cdots,w_{t-1}$ are the weights and $b$ is the bias. 

Widely used activation functions include logistic, hyperbolic tangent, and rectified linear unit \cite{bircanouglu2018comparison}.
\begin{align*}
	f(.)= \frac{1}{1+e^{(-.)}}.
\end{align*}

\begin{align*}
    \tanh(.)=\frac{e^{(.)}-e^{(-.)}}{e^{(.)}+e^{(-.)}}.
\end{align*}

\begin{align*}
	R(.)= \max(0,.).
\end{align*}

\subsubsection{Gaussian Processes}
Gaussian processes can be thought of as infinite dimensional Gaussian distributions \cite{rasmussen2003gaussian}. Assume that we are given a function:
\begin{eqnarray}
	f(.)\sim GP(m(.),k(.,.'))
\end{eqnarray}
where $m(.)$ is the mean function and $k(.,.')$ is the covariance function. An example of covariance function is the squared exponential kernel function:
\begin{eqnarray}\label{tttt}
	k(.,.')=\sigma\exp(-\frac{1}{l^2}(.-.')^2)
\end{eqnarray}
with the hyperparemeters $\sigma$ and $l$, the kernel variance and  the length scale respectively.
The observation model is given as follows:
\begin{eqnarray}
    Y = f + \epsilon
\end{eqnarray}
where $\epsilon \sim N_n(0,\sigma^2 I)$. The Gram matrix is given by its entries $K_{i,j}=k(._i,._j)$ and the prediction of is as follows:
\begin{eqnarray}
    f(.)|Y \sim GP(\mu', k' )
\end{eqnarray}
with $\mu'=k^T(.)(K+\sigma^2I)^{-1})Y$ and $k'=k(.,.)-k(.)^T(K+\sigma I)^{-1}    k(.) .$


\section{Results and Discussion}
\subsection{Trend analysis}
\begin{figure}[ht!]
    \centering
		\caption{Calendar Plots of Daily PM2.5 Concentration at Gitega}
		\includegraphics[width=1.02\linewidth]{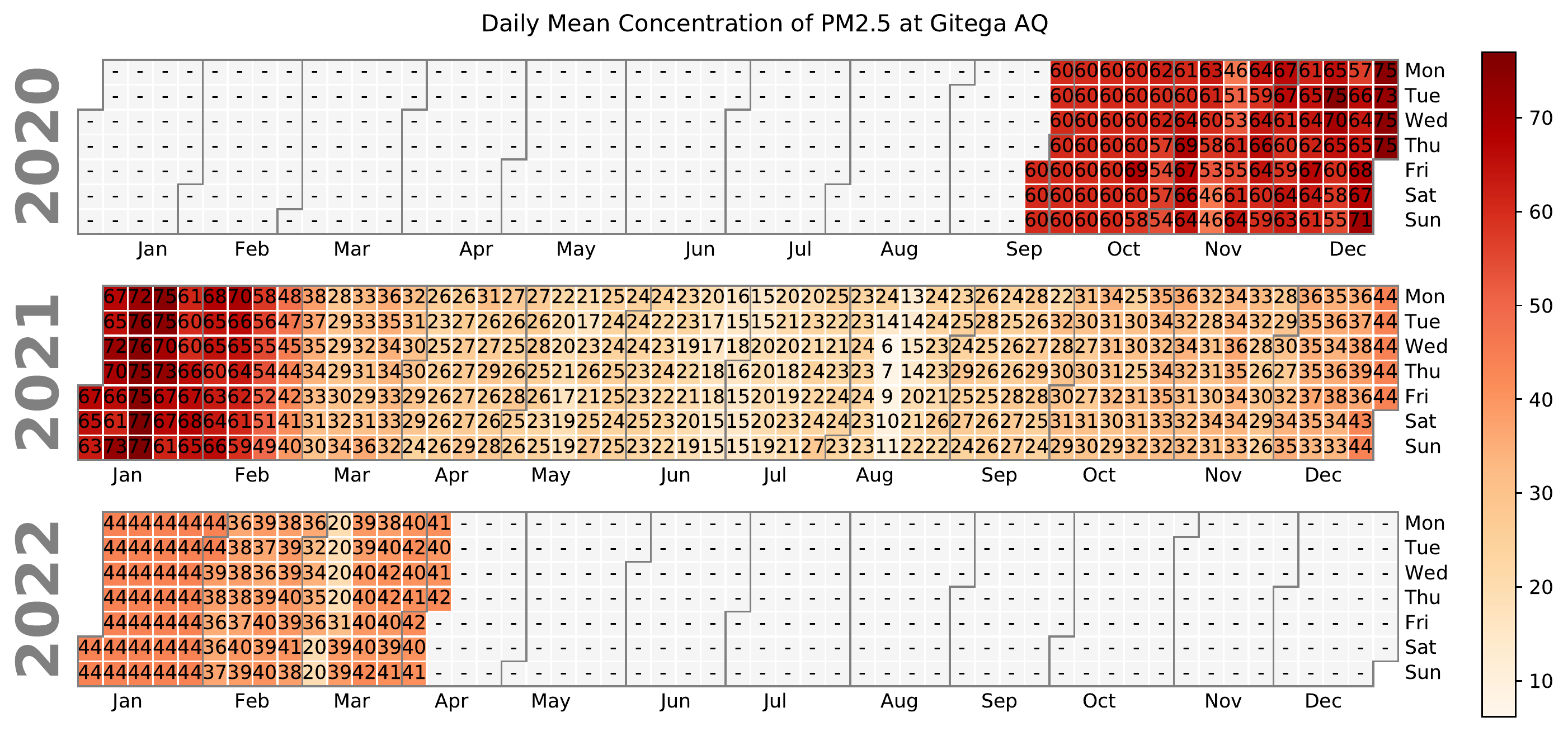}  
		\label{fig:sub-first_calendar}
\end{figure}

\begin{figure}[ht!]
    \centering
		\caption{Long dry season (June to August)}
		\includegraphics[width=1.01\linewidth]{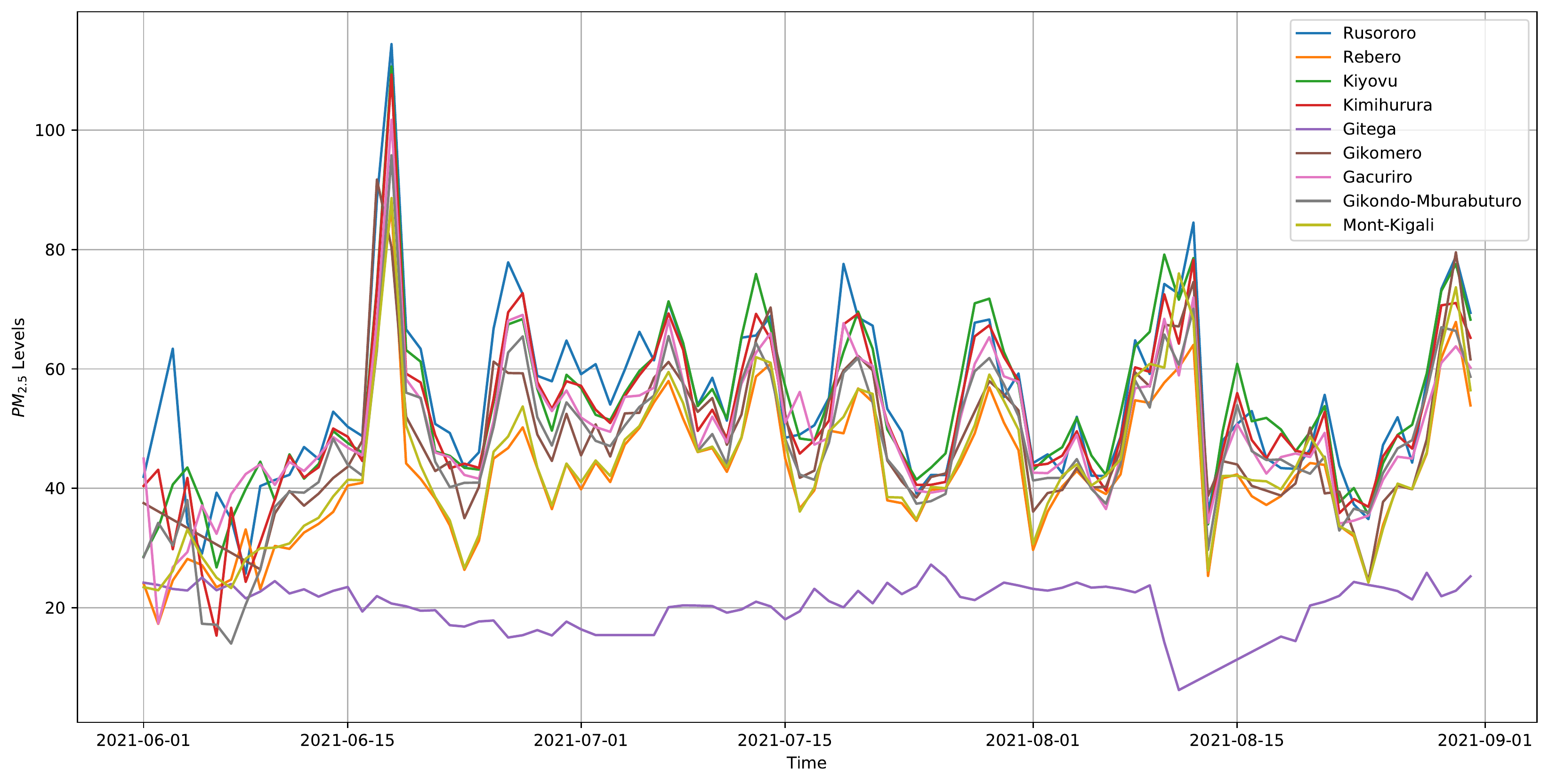}  
		\label{fig:sub-first_Seasonal}
\end{figure}

The hourly analysis was carried out to determine which hours had high PM2.5 emissions based on the sources of the emissions and other related activities that significantly increased the rate of emissions at each air monitoring station. In Rwanda, both public and private transports are used in towns when moving from home to office and it is observed that PM2.5 concentrations are higher in the morning hours when people are going to work and start to decrease during mid-day hours. The indoor activities are also sources of particulate matter: cooking and fuel-burning mainly. These additional pollution sources  are also impacting the overall pollution levels in the remaining hours. The figures below show the box-plots drawn to examine the hourly mean of PM2.5 concentrations measured by each station around Kigali. They enabled us to examine the minimum, first and third quartile, inter-quartile range, median, and maximum values observed when the air quality monitoring sensors were recording the real-time PM2.5 concentrations. It was observed that after the morning hours, pollution levels begin to decline during mid-day hours and then rise again in the afternoon when people leave office. In the evening, pollution from biomass burning as a source of energy is high in various areas of Kigali. Gikomero, Gacuriro, Rusororo, and Rebero air monitoring stations have the highest hourly mean PM2.5 concentrations of $\SI{214.516}{\micro\gram}/m^3$, $\SI{218.756}{\micro\gram}/m^3$, $\SI{196.958}{\micro\gram}/m^3$, and $\SI{189.716}{\micro\gram}/m^3$ respectively. The highest median hourly $PM2.5$ concentrations are recorded at Rusororo, Kimihurura, Kiyovu, and Gikondo Mburabuturo, with values of $\SI{46.831}{\micro\gram}/m^3$, $\SI{44.952}{\micro\gram}/m^3$, $\SI{44.695}{\micro\gram}/m^3$, and $\SI{43.228}{\micro\gram}/m^3$ respectively. This indicates that emissions from traffic vehicles and biomass combustion have influenced the rate of air pollution and increased the country's average, which has exceeded $\SI{15}{\micro\gram}/m^3$ as a 24-hour mean, as recommended by WHO guidelines for September 2021.
\begin{figure}[ht!]
    \centering
\caption{Hourly PM2.5 concentrations at Gitega air monitoring station}
		\includegraphics[width=1.092\linewidth]{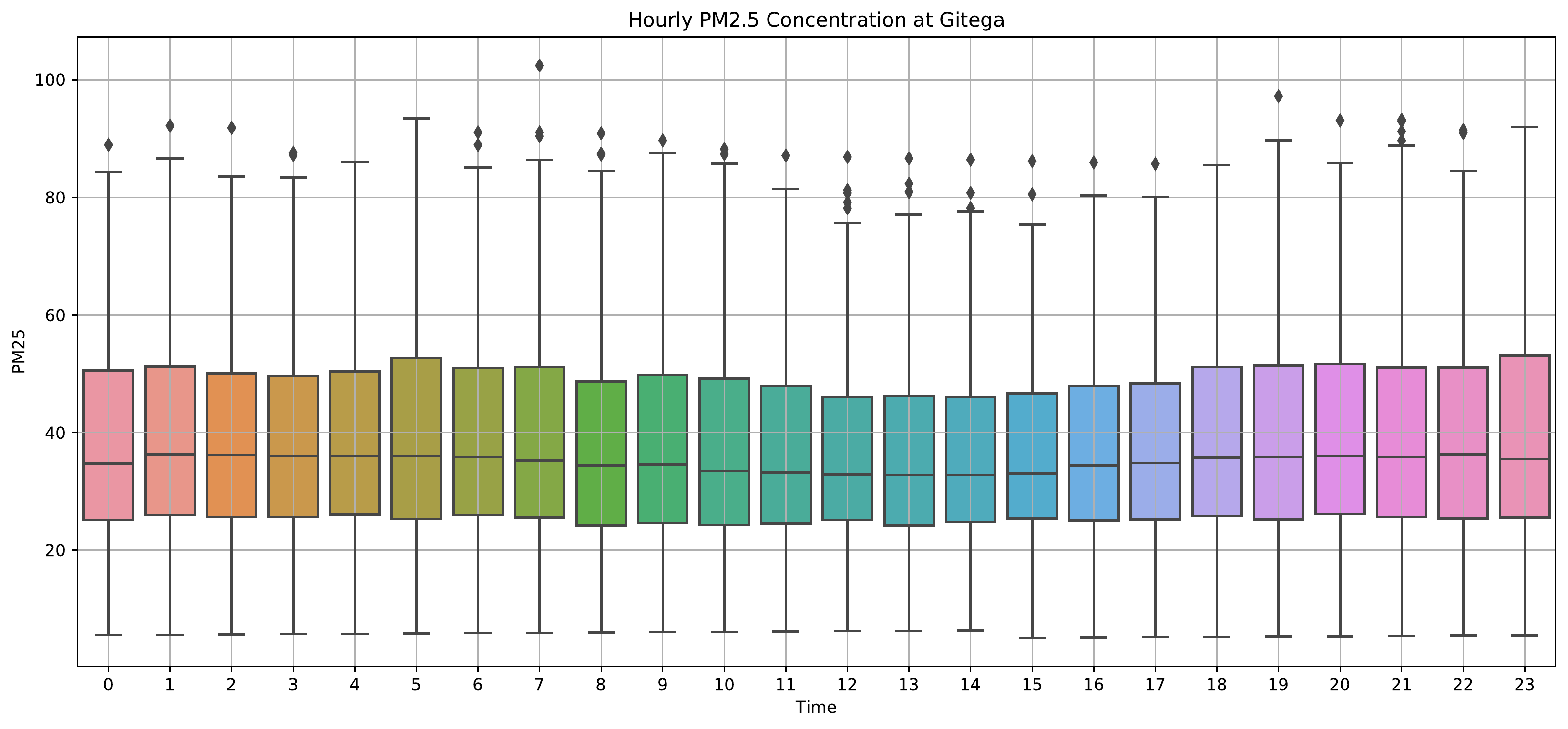}  
		\label{fig:sub-first_Hourly}
\end{figure}
\FloatBarrier
\begin{figure}[ht!]
    \centering
		\caption{Hourly PM2.5 concentrations at Gacuriro air monitoring station}
		\includegraphics[width=1.092\linewidth]{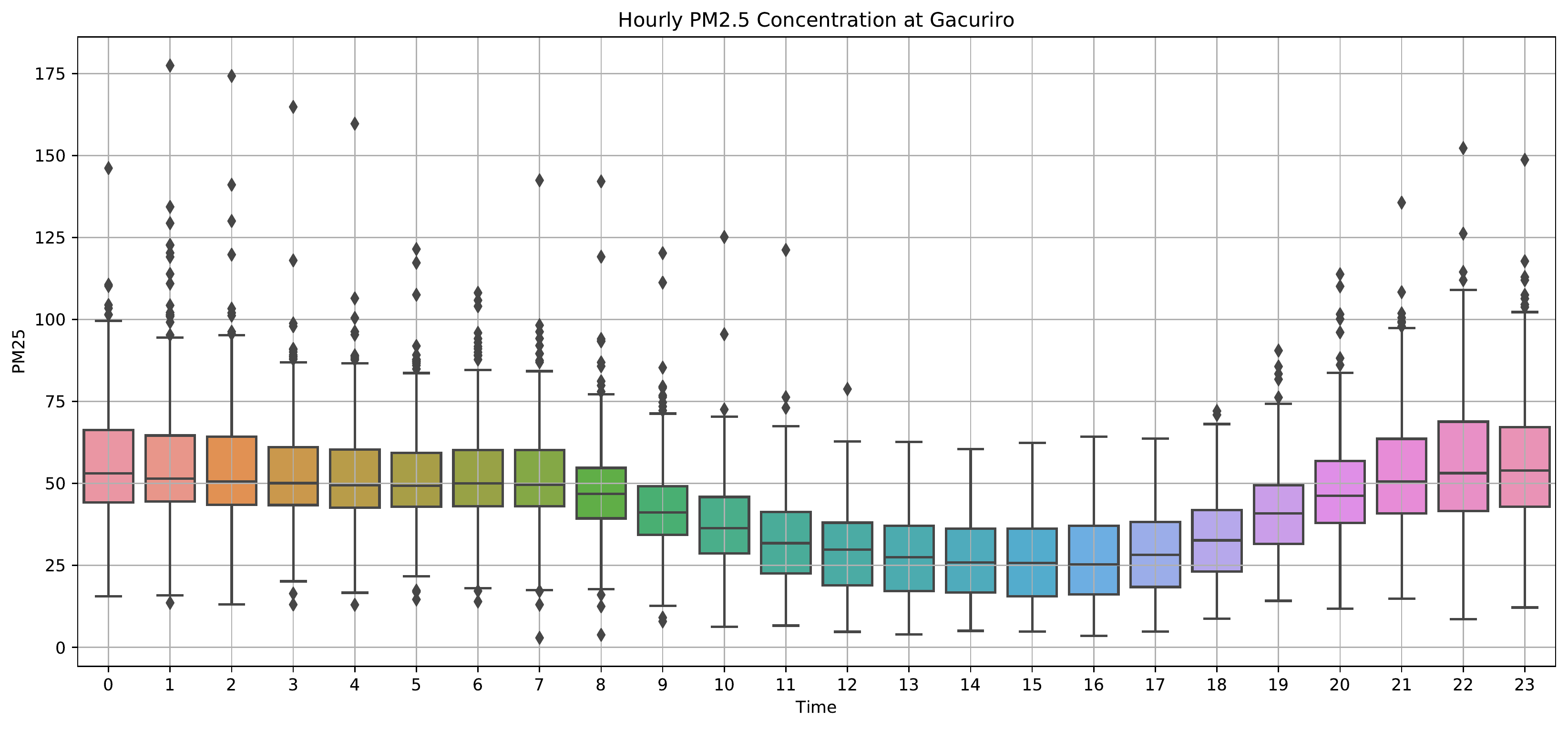}  
		\label{fig:sub-second_Hourly}
\end{figure}
\FloatBarrier
\begin{figure}[ht!]
    \centering
		\caption{Hourly PM2.5 concentrations at Rebero air monitoring station }
		\includegraphics[width=1.092\linewidth]{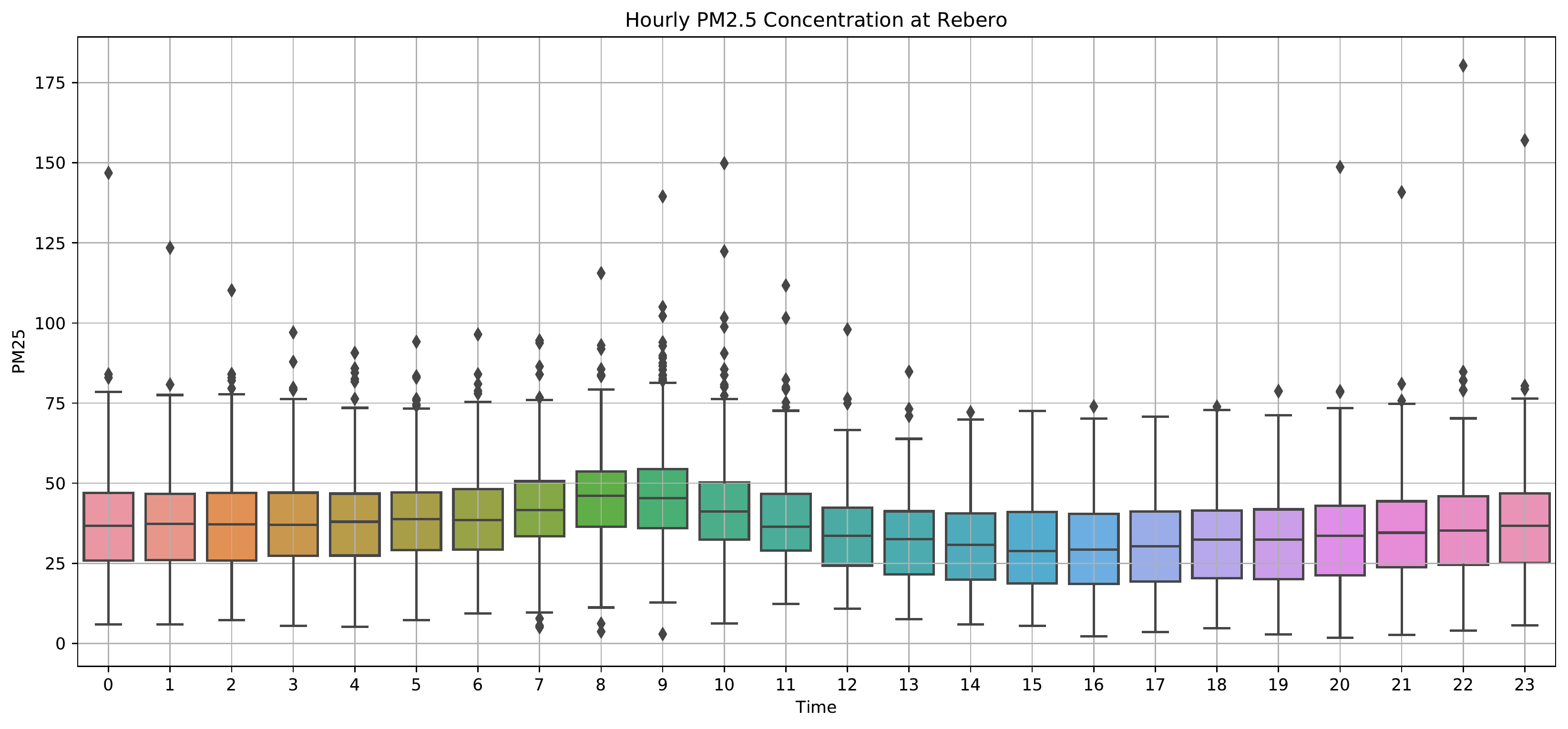}  
		\label{fig:sub-third_Hourly}
\end{figure}
\FloatBarrier
\begin{figure}[ht!]
    \centering
		\caption{Hourly PM2.5 concentrations at Kimihurura air monitoring station }
		\includegraphics[width=1.092\linewidth]{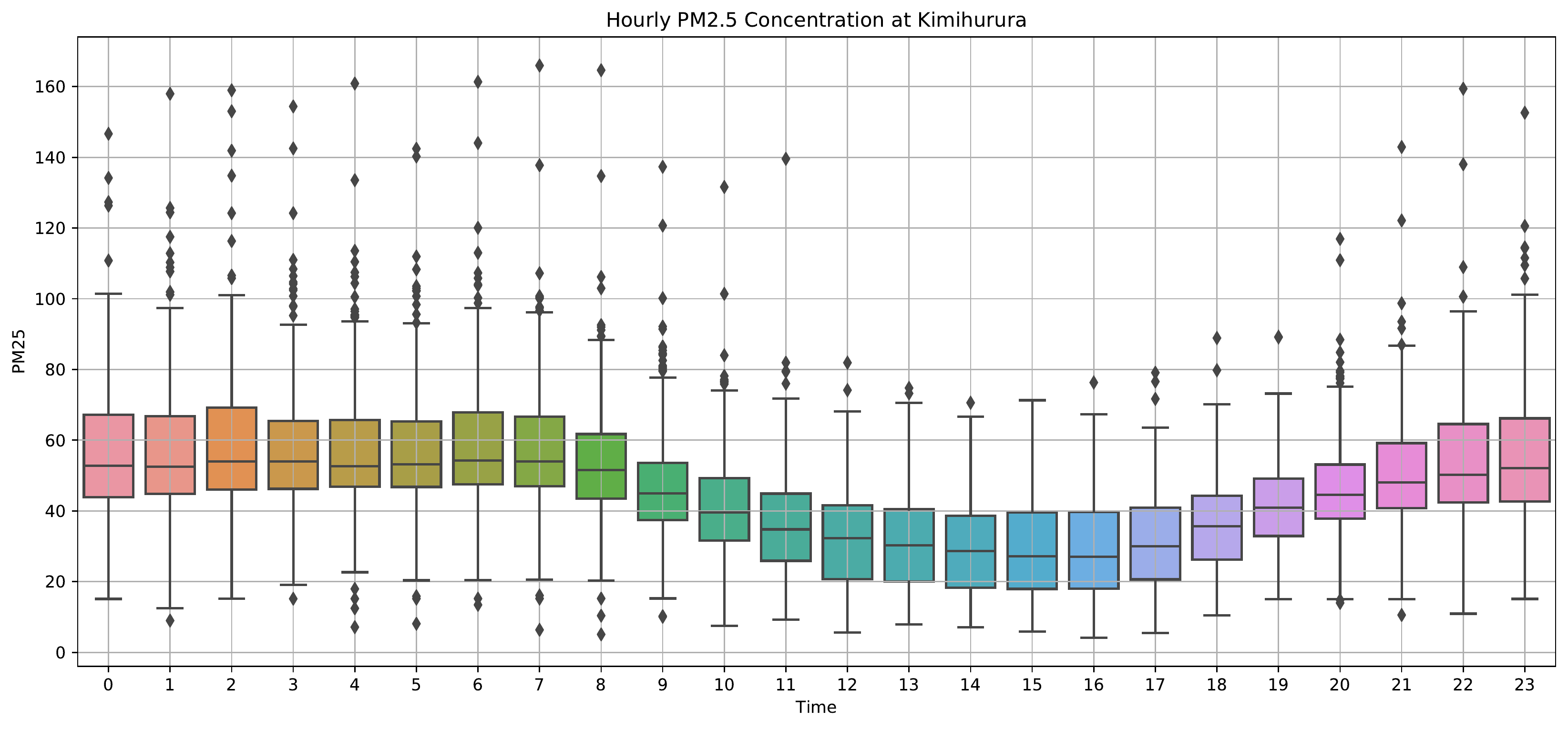}  
		\label{fig:sub-fourth_Hourly}
\end{figure}
\FloatBarrier
\begin{figure}[ht!]
    \centering
		\caption{Hourly PM2.5 concentrations at Gikondo air monitoring station }
		\includegraphics[width=1.092\linewidth]{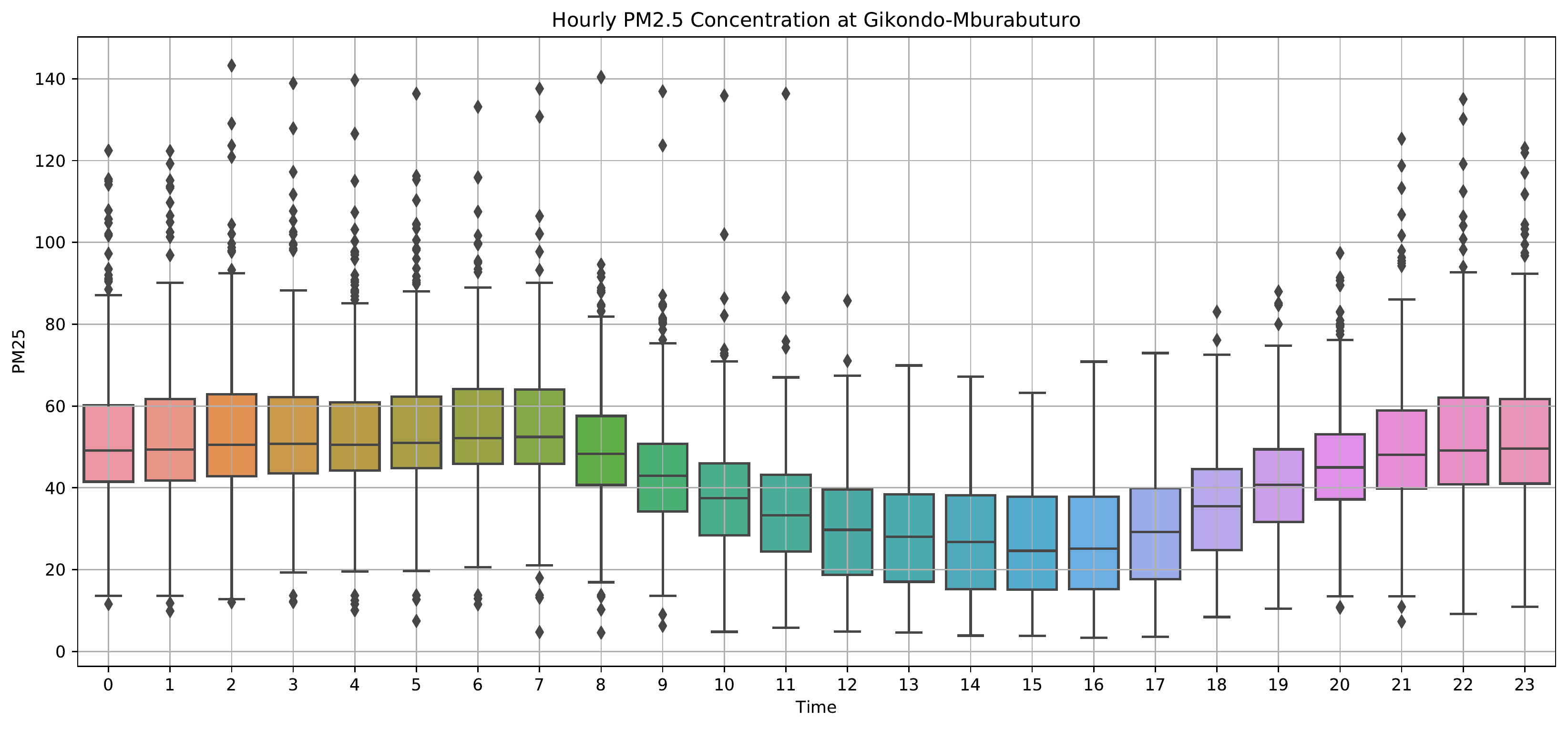}  
		\label{fig:sub-fifth_Hourly}
\end{figure}
\FloatBarrier
\begin{figure}[ht!]
    \centering
		\caption{Hourly PM2.5 concentrations at Kiyovu air monitoring station }
		\includegraphics[width=1.092\linewidth]{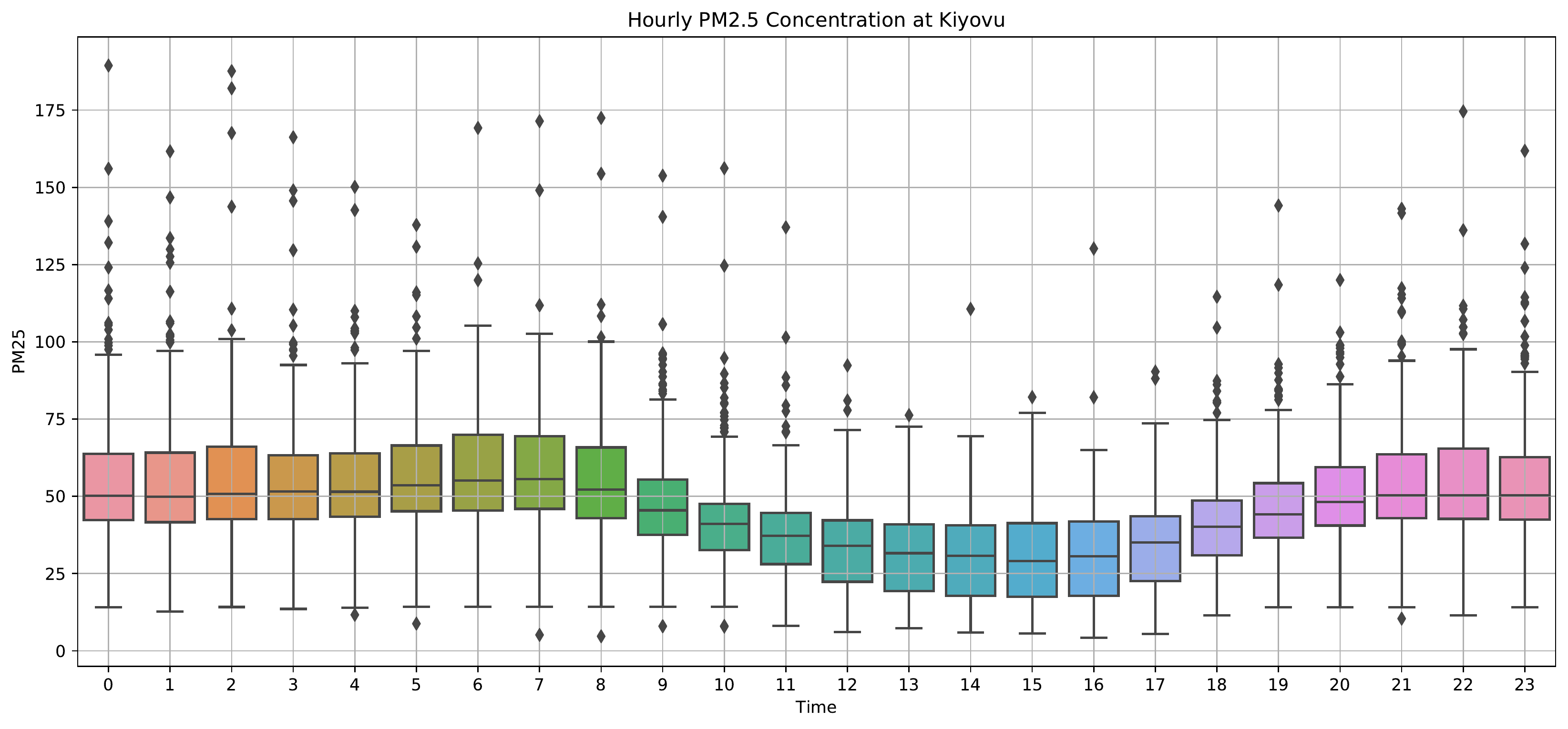}  
		\label{fig:sub-sixth_Hourly}
\end{figure}
\FloatBarrier
\begin{figure}[ht!]
    \centering
		\caption{Hourly PM2.5 concentrations at Mount Kigali air monitoring station }
		\includegraphics[width=1.092\linewidth]{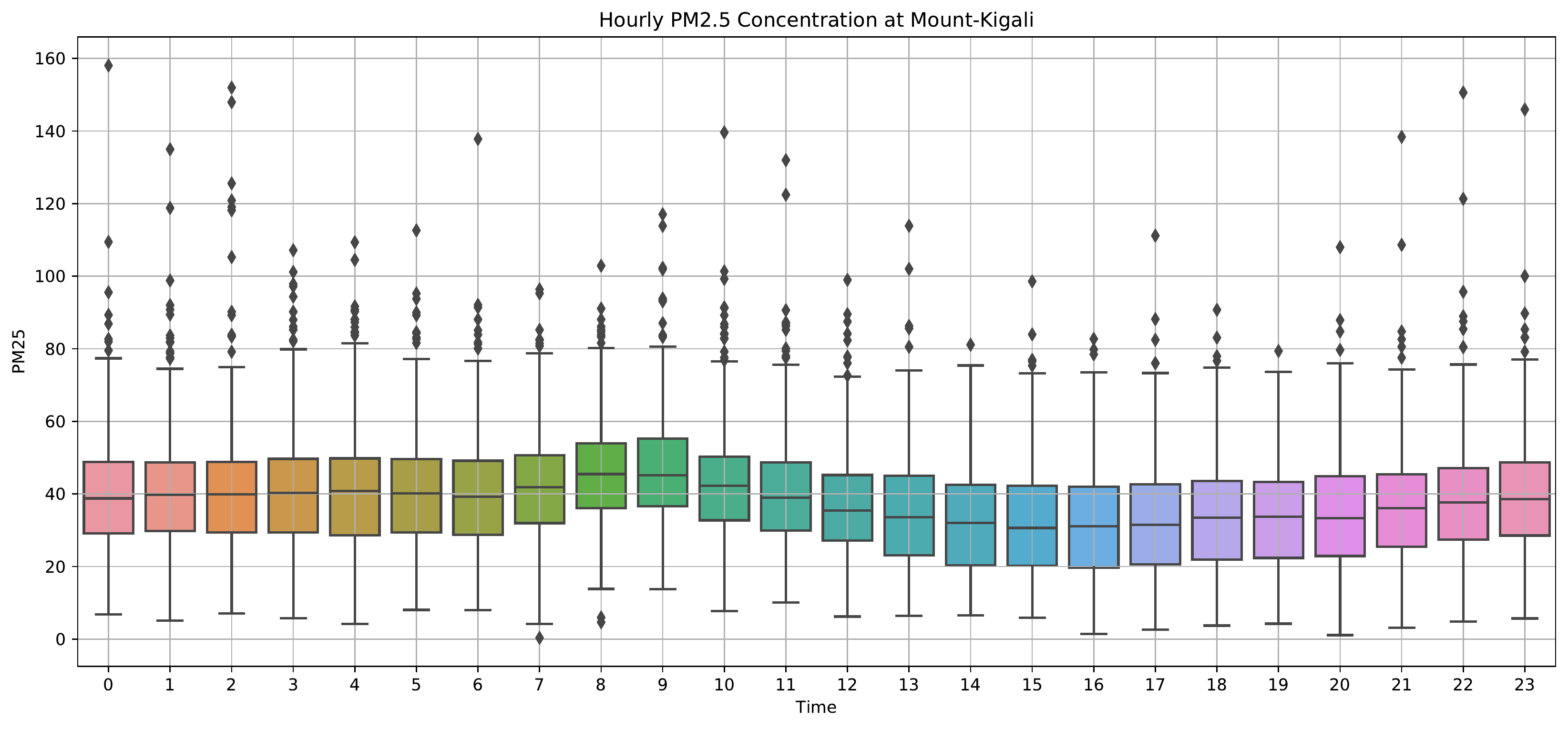} \label{fig:sub-seventh_Hourly}
\end{figure}
\FloatBarrier
\begin{figure}[ht!]
    \centering
		\caption{Hourly PM2.5 concentrations at Rusororo air monitoring station }
		\includegraphics[width=1.092\linewidth]{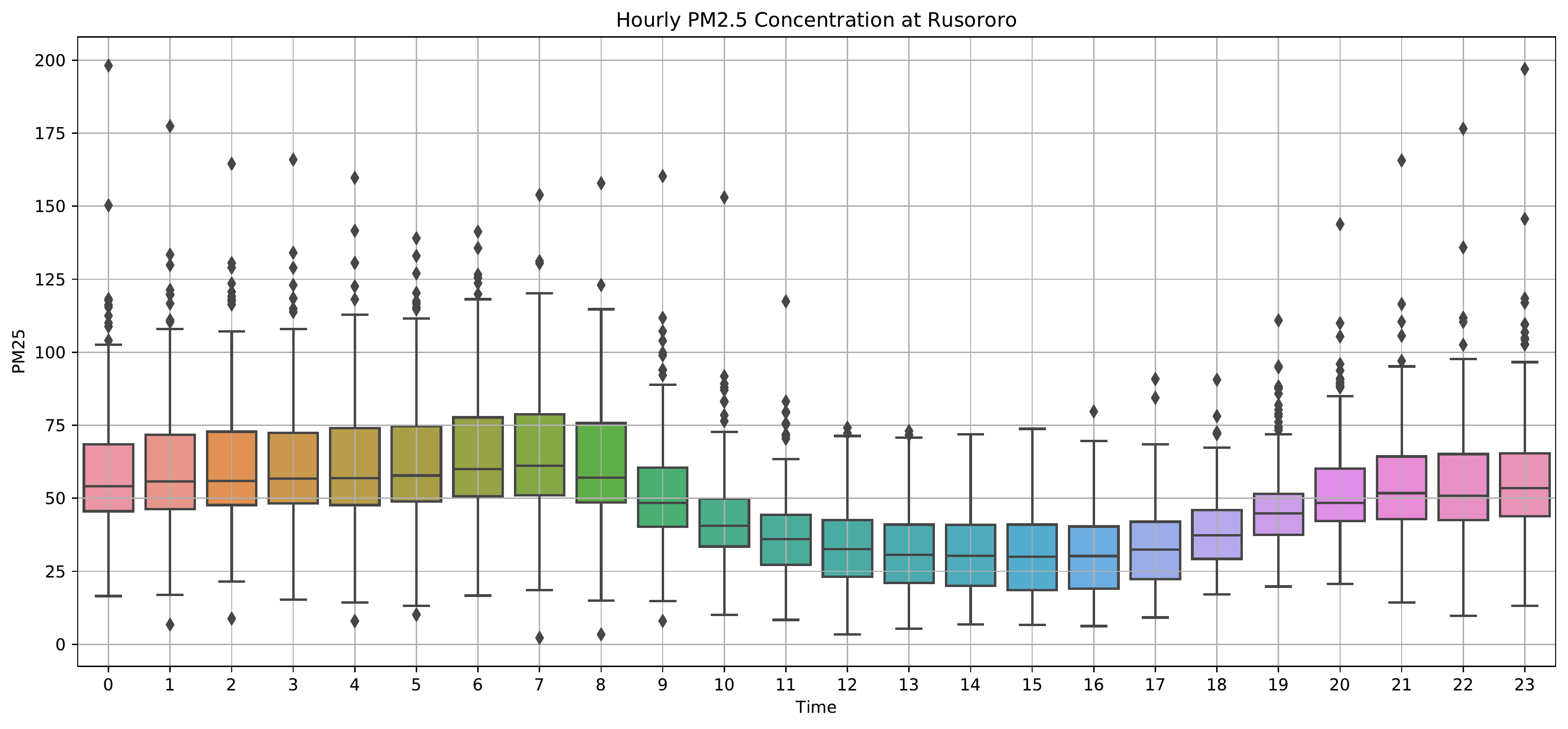}  
		\label{fig:sub-eighth_Hourly}
\end{figure}
\FloatBarrier
\begin{figure}[ht!]
    \centering
		\caption{Hourly PM2.5 concentrations at Gikomero air monitoring station }
		\includegraphics[width=1.092\linewidth]{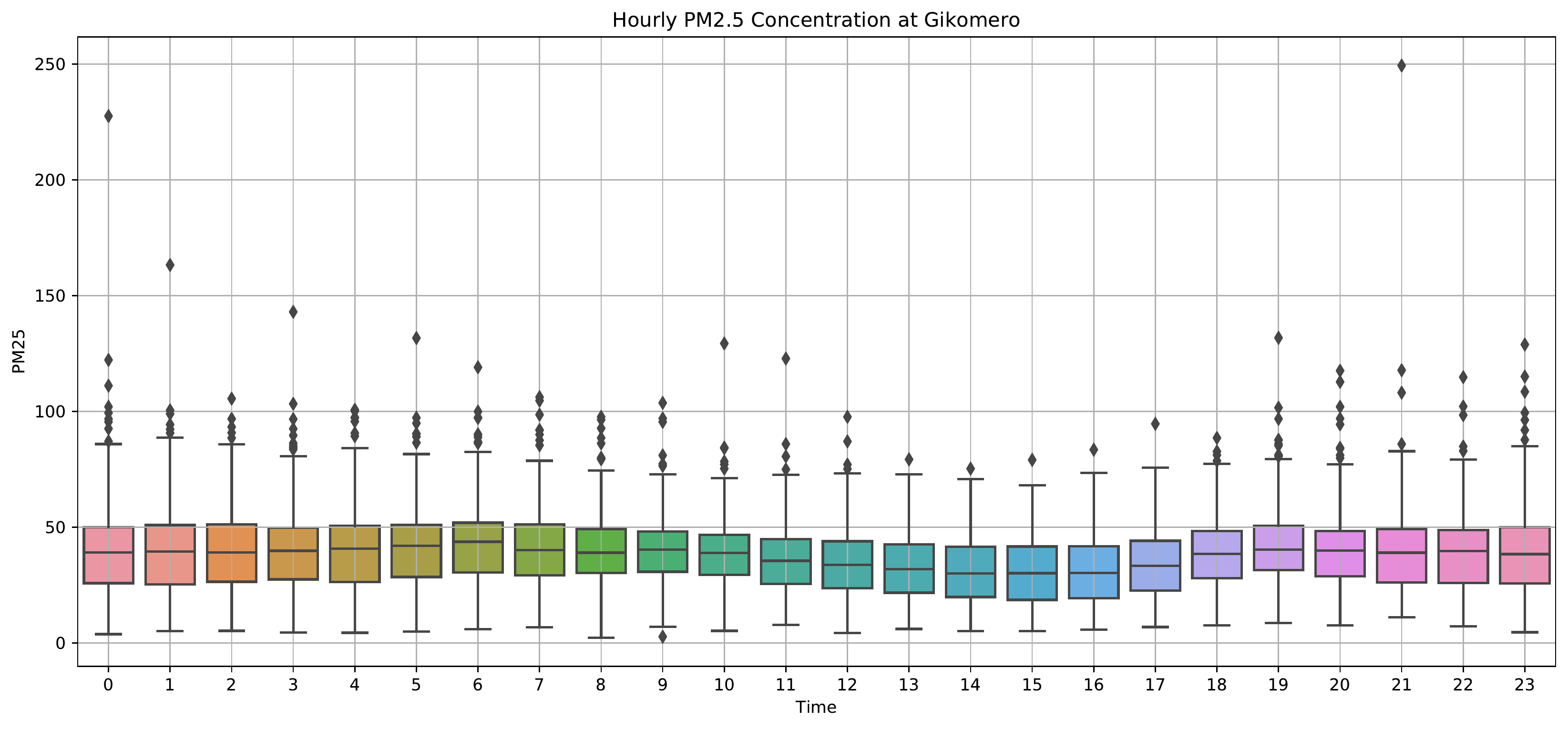}  
		\label{fig:sub-ninth_Hourly}
\end{figure}
\FloatBarrier
The daily mean of PM2.5 concentrations at each station was examined to determine which days of the week had the highest rate of emissions in comparison to others. This analysis, which uses box plots, compares emissions during working days and weekends, as well as determines which days have the highest emissions. It appears that weekdays have the highest levels compared to weekends. Gikomero, Mount-Kigali, and Rusororo monitoring stations revealed that emissions are higher on Thursday than on other days, whereas Wednesday was found to be the day with the highest emissions at Gacuriro, Kiyovu, Rebero, and Kimihurura. The high pollution levels on those days may be  attributable to traffic emissions, as many people in Kigali drive to work and on weekends traffic is limited. Although on weekends traffic related pollution is reduced, the emission from industrial sites and biomass combustion remain constant. 
\begin{figure}[ht!]
    \centering
		\caption{Daily PM2.5 concentrations at Gitega air monitoring station }
		\includegraphics[width=1.092\linewidth]{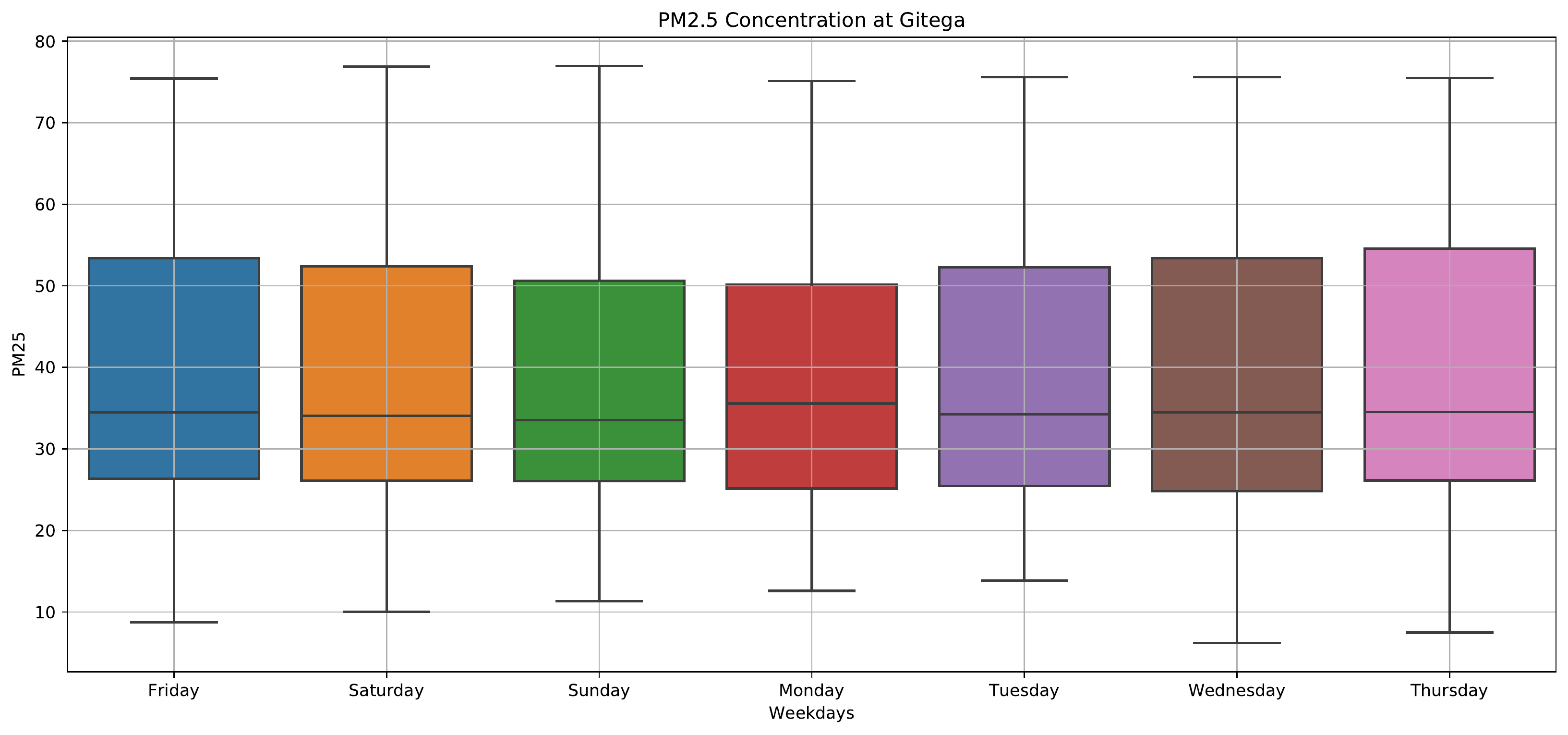}  
		\label{fig:sub-first_Daily}
\end{figure}
\FloatBarrier
\begin{figure}[ht!]
    \centering
		\caption{Daily PM2.5 concentrations at Gacuriro air monitoring station }
		\includegraphics[width=1.092\linewidth]{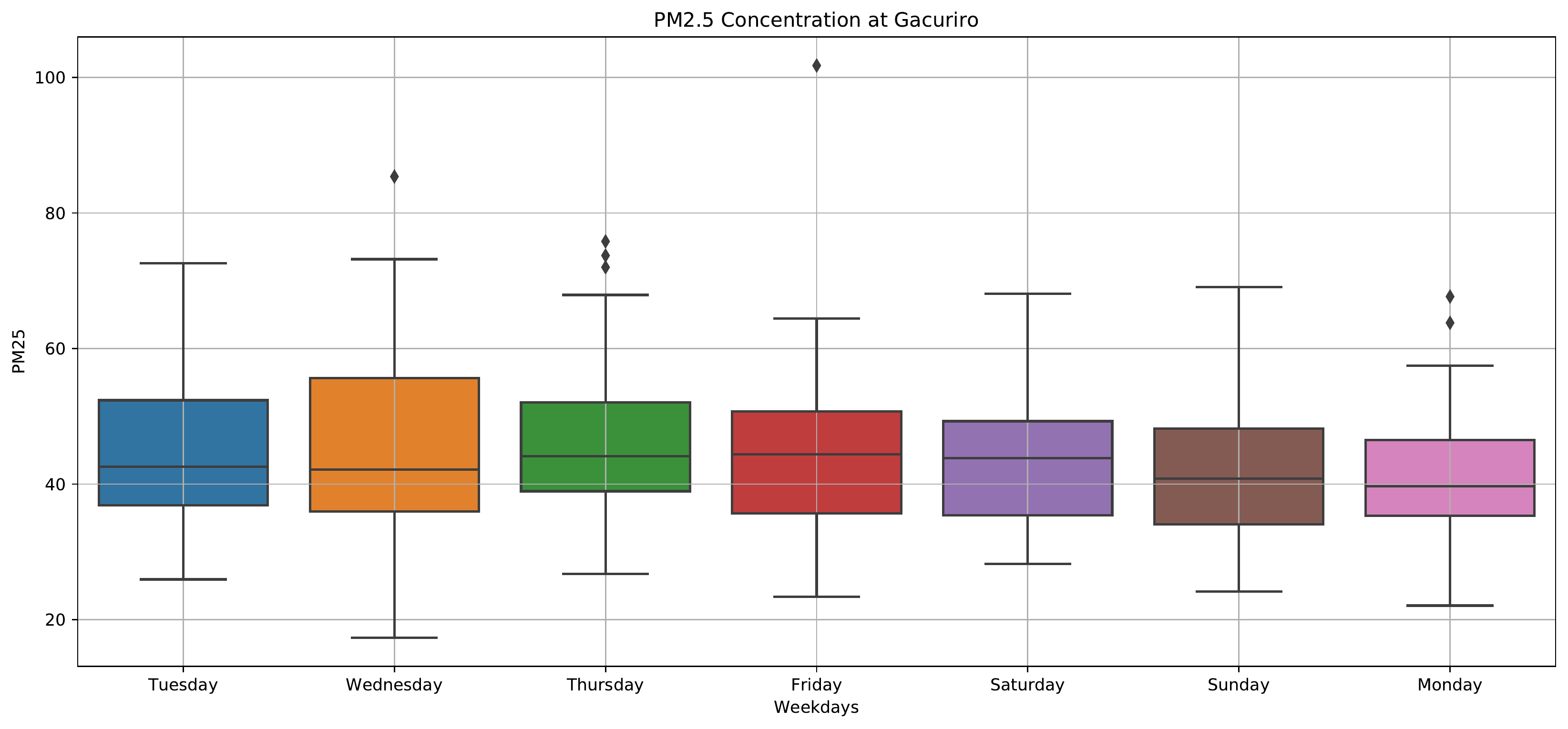}  
		\label{fig:sub-second_Daily}
\end{figure}
\FloatBarrier
\begin{figure}[ht!]
    \centering
		\caption{Daily PM2.5 concentrations at Rebero air monitoring station }
		\includegraphics[width=1.092\linewidth]{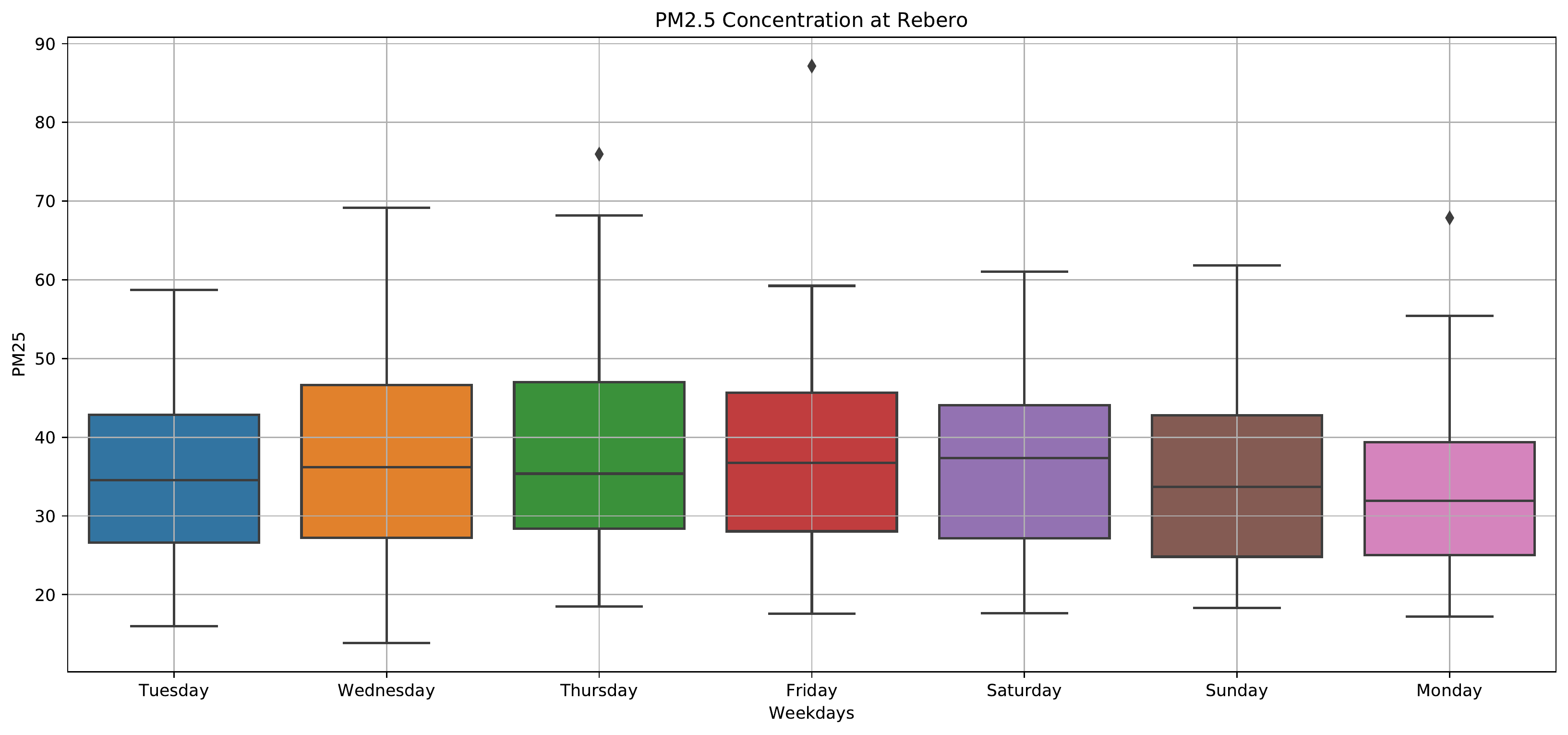}  
		\label{fig:sub-third_Daily}
\end{figure}
\FloatBarrier
\begin{figure}[ht!]
    \centering
		\caption{Daily PM2.5 concentrations at Kimihurura air monitoring station }
		\includegraphics[width=1.092\linewidth]{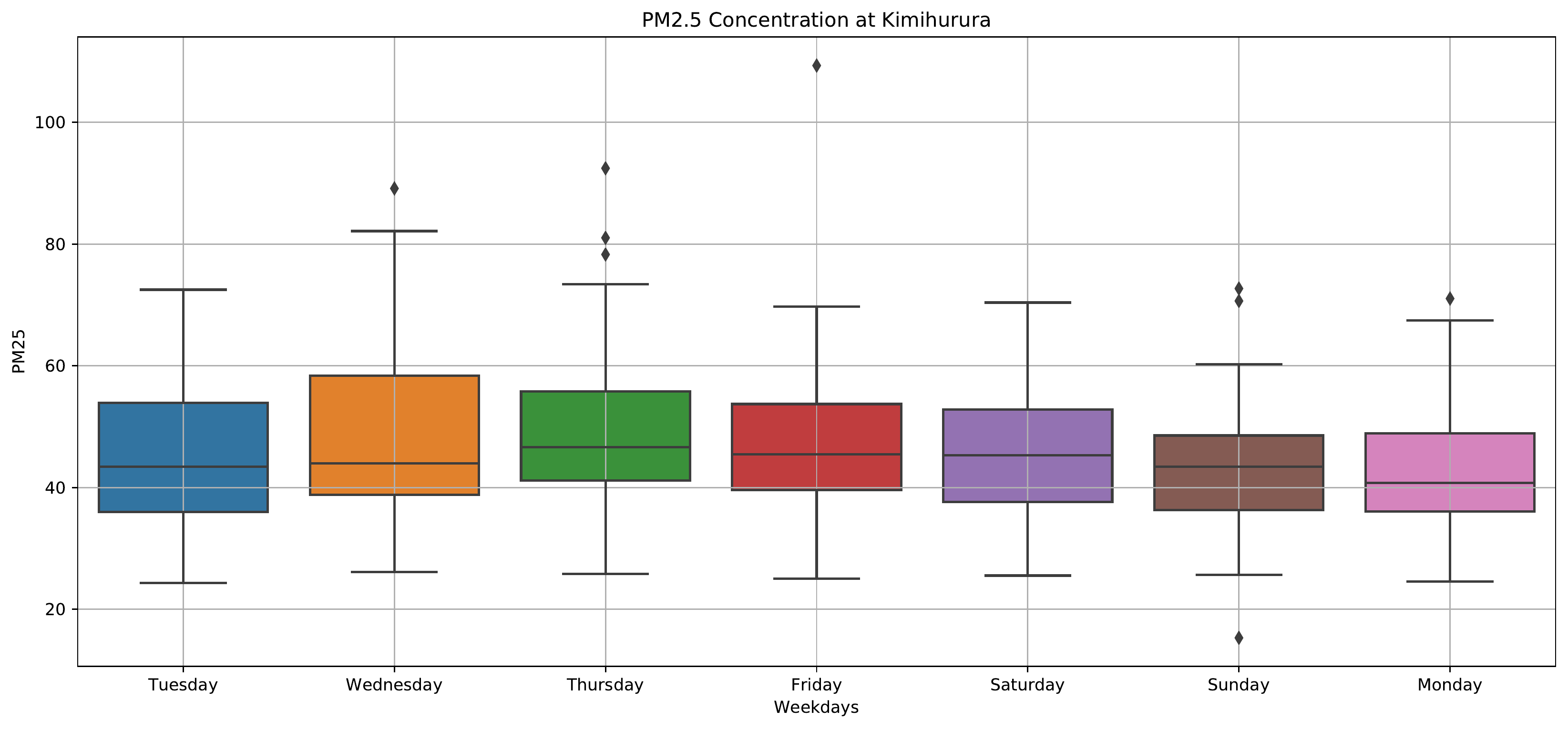}  
		\label{fig:sub-fourth_Daily}
\end{figure}
\FloatBarrier
\begin{figure}[ht!]
    \centering
		\caption{Daily PM2.5 concentrations at Gikondo air monitoring station }
		\includegraphics[width=1.092\linewidth]{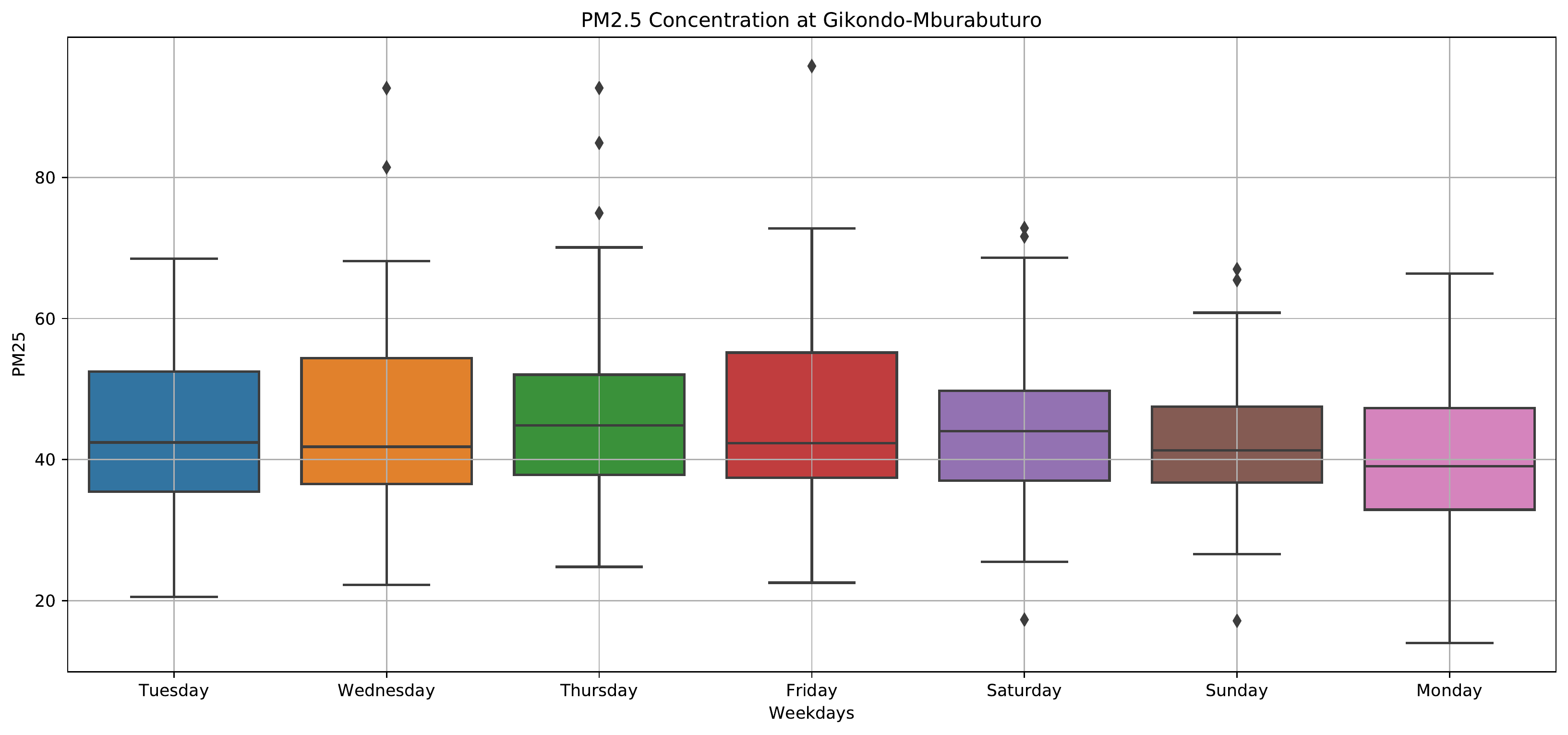}  
		\label{fig:sub-fifth_Daily}
\end{figure}
\FloatBarrier
\begin{figure}[ht!]
    \centering
		\caption{Daily PM2.5 concentrations at Kiyovu air monitoring station }
		\includegraphics[width=1.092\linewidth]{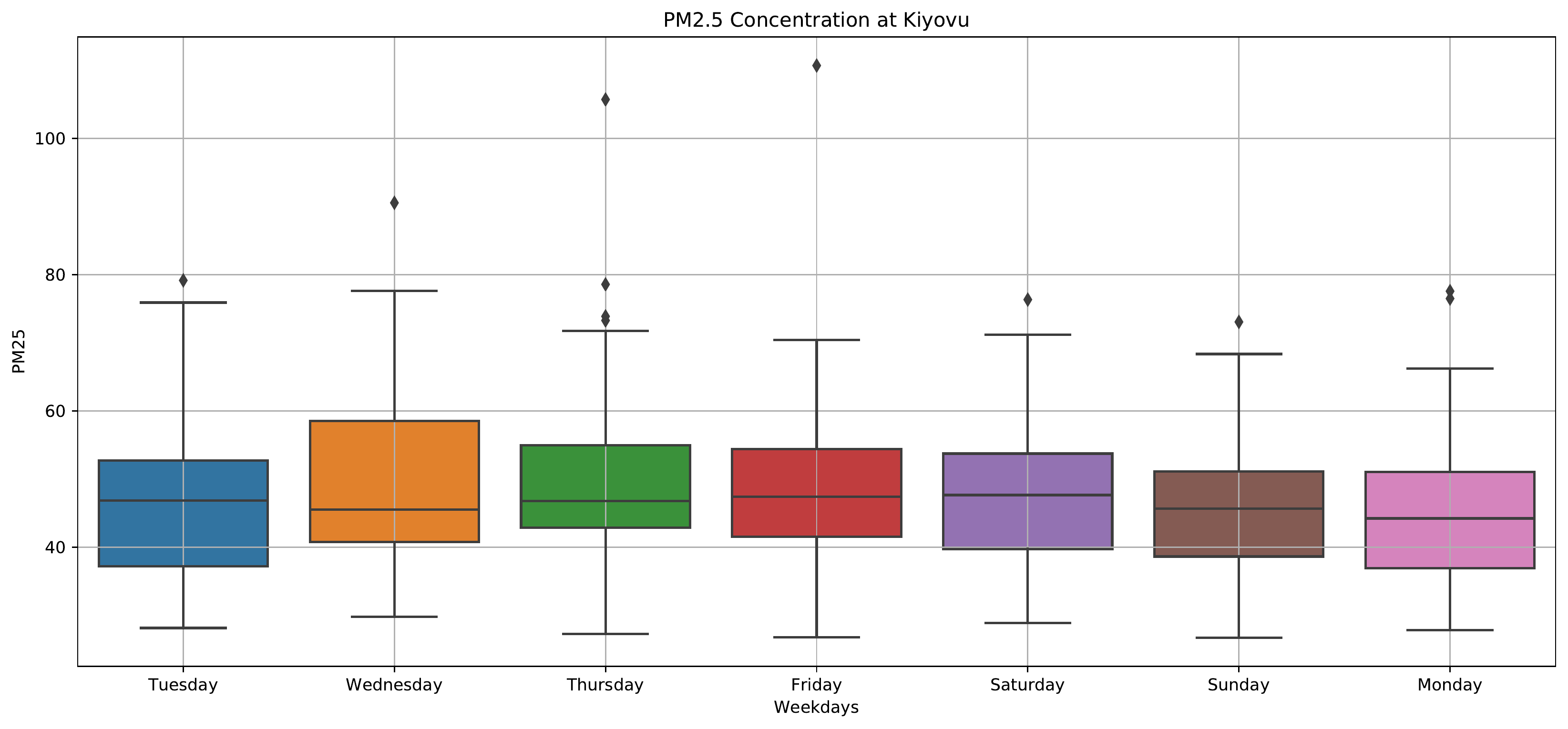}  
		\label{fig:sub-sixth_Daily}
\end{figure}
\FloatBarrier
\begin{figure}[ht!]
    \centering
		\caption{Daily PM2.5 concentrations at Mount Kigali air monitoring station }
		\includegraphics[width=1.02\linewidth]{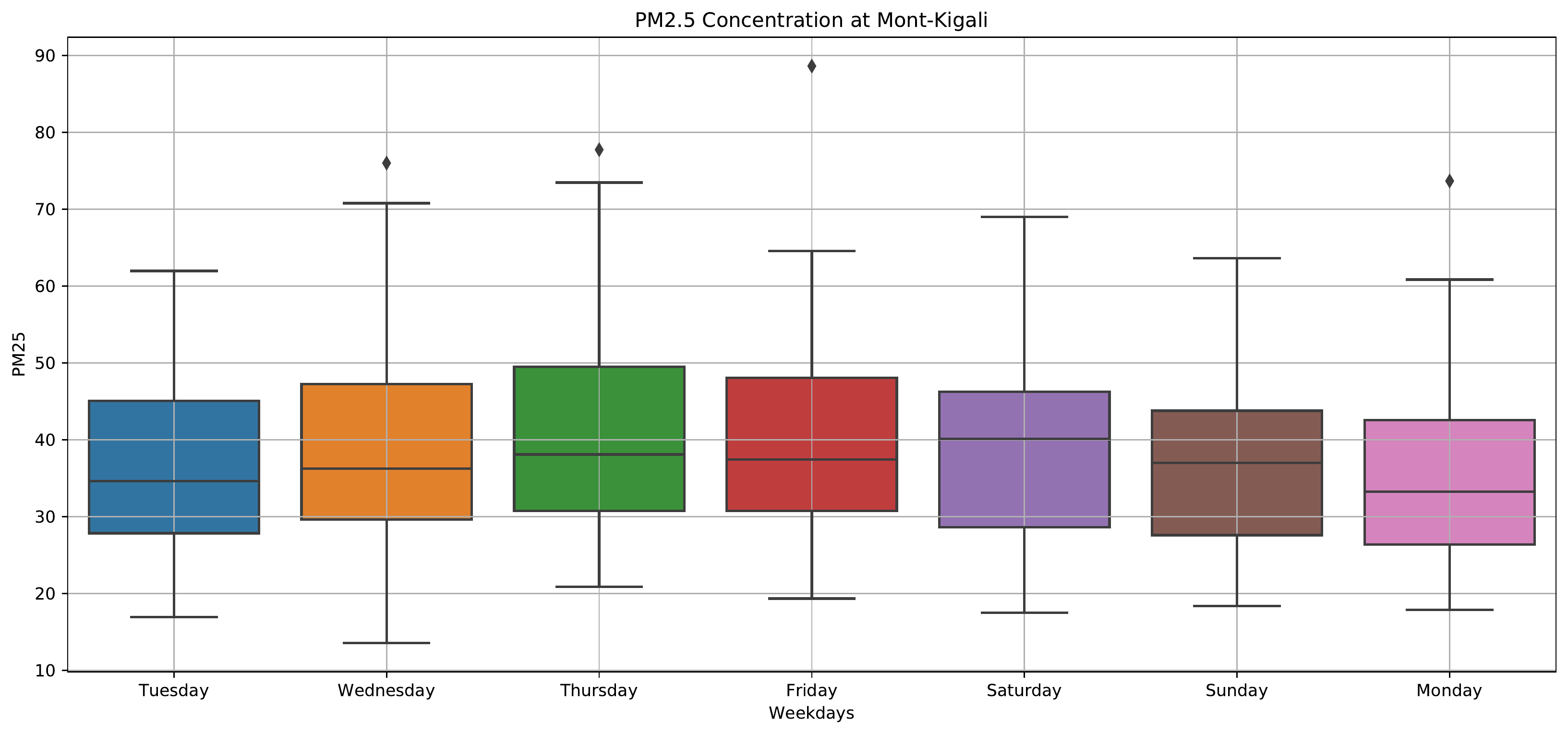} \label{fig:sub-seventh_Daily}
\end{figure}
\FloatBarrier
\begin{figure}[ht!]
    \centering
		\caption{Daily PM2.5 concentrations at Rusororo air monitoring station }
		\includegraphics[width=1.02\linewidth]{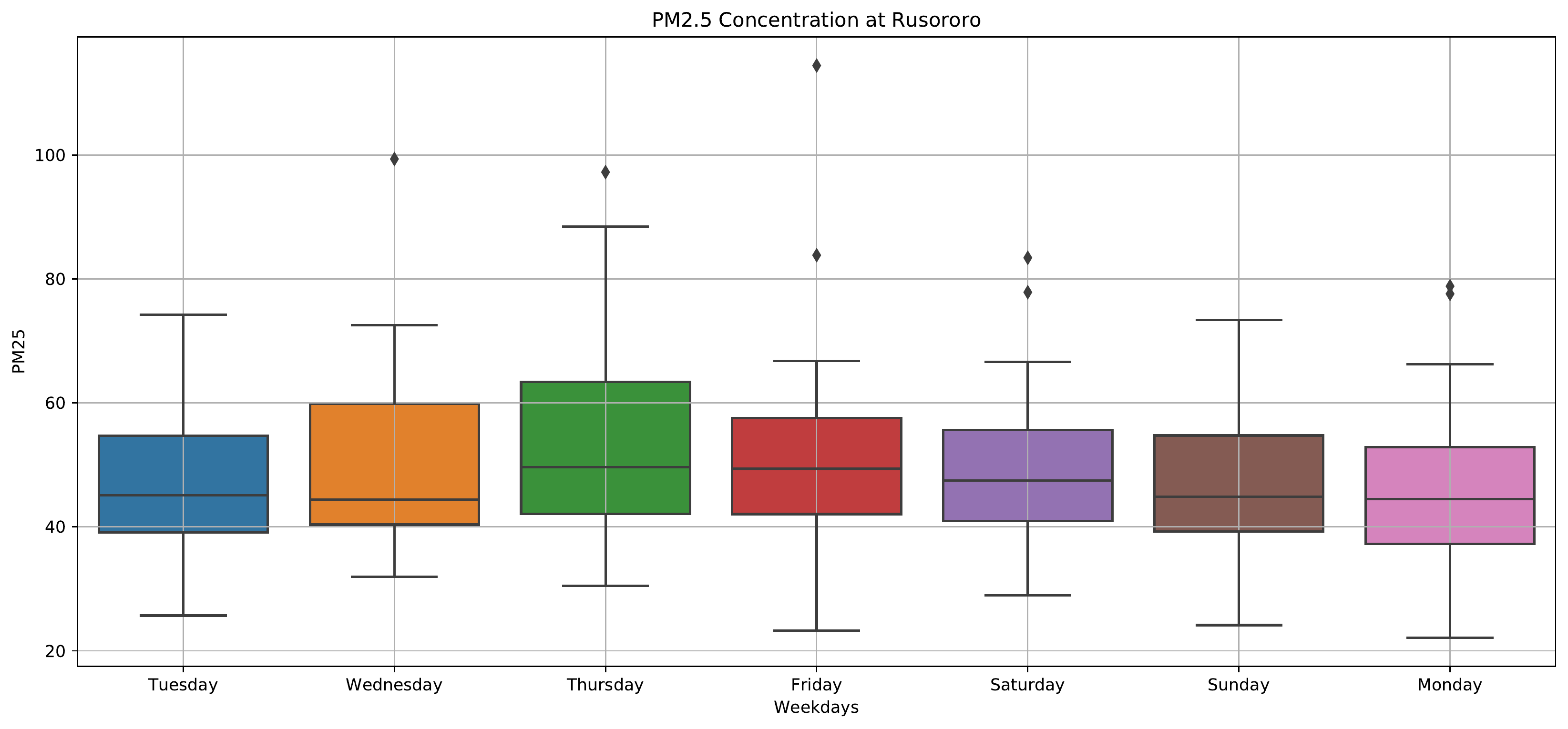}  
		\label{fig:sub-eighth_Daily}
\end{figure}
\FloatBarrier
\begin{figure}[ht!]
    \centering
		\caption{Daily PM2.5 concentrations at Gikomero air monitoring station }
		\includegraphics[width=1.092\linewidth]{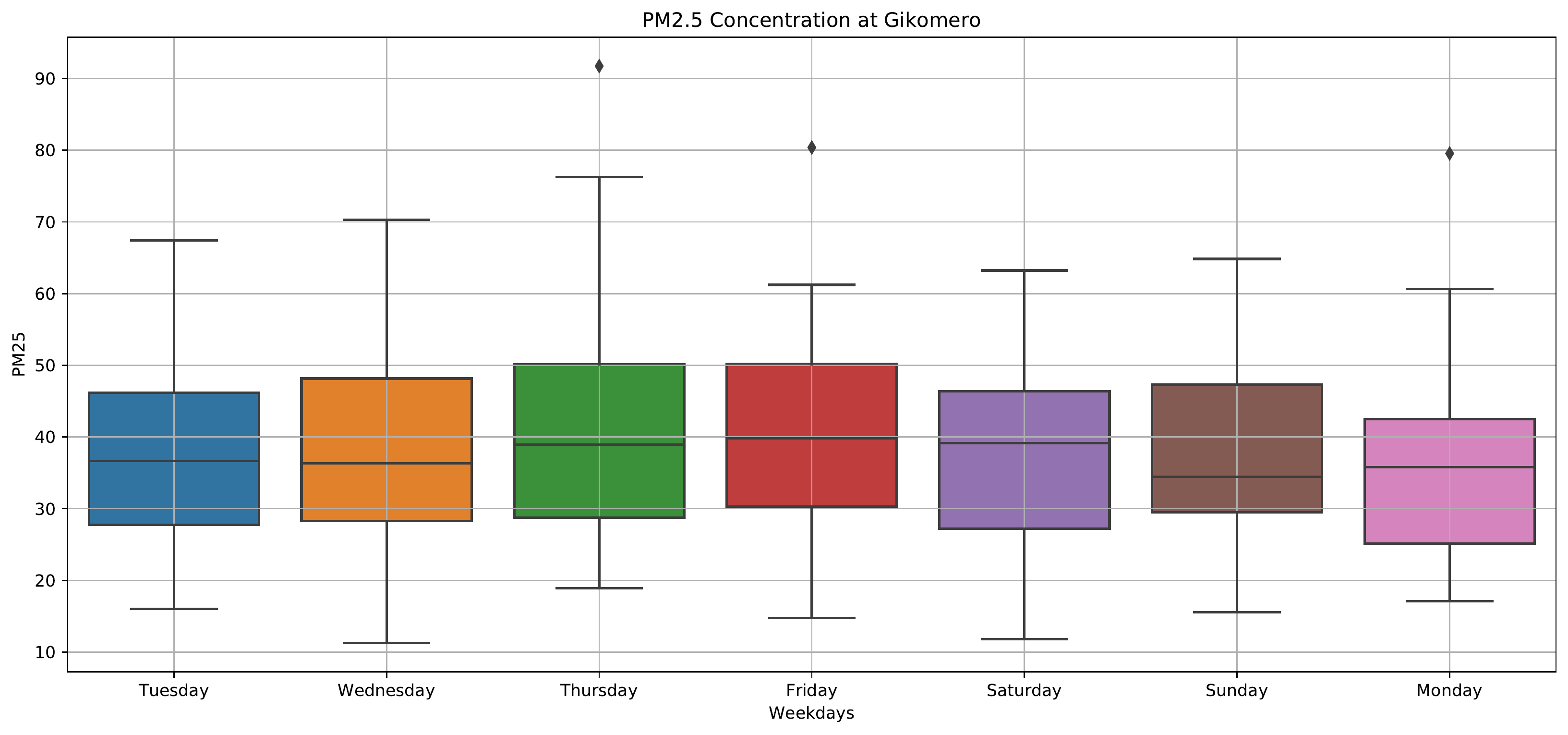}  
		\label{fig:sub-ninth_Daily}
\end{figure}
\FloatBarrier
The daily mean of PM2.5 concentrations obtained at each air monitoring station has been connected to the daily calendar to track variations over time. The calendar analysis was useful in determining which days and months had high and low PM2.5 concentrations. In the months of September, October, November, and December $2020$, the PM2.5 concentration at the Gitega monitoring station (\ref{fig:sub-first_calendar}) was particularly high. It was also high at the start of $2021$, but it began to decline by the end of February $2021$. Daily PM2.5 concentrations were higher at all stations in the last two months of the short wet season (October and November $2021$) and the first month of the short dry season (December $2021$) than on other days. Hence, precipitation acts as a dust collector, the reduction in PM2.5 concentrations in those months was related to the amount of precipitation observed during the short wet season $2021$ and December $2021$. 
\FloatBarrier
\begin{figure}[ht!]
    \centering
		\caption{Calendar Plots of Daily PM2.5 Concentration  at Gacuriro}
		\includegraphics[width=1.02\linewidth]{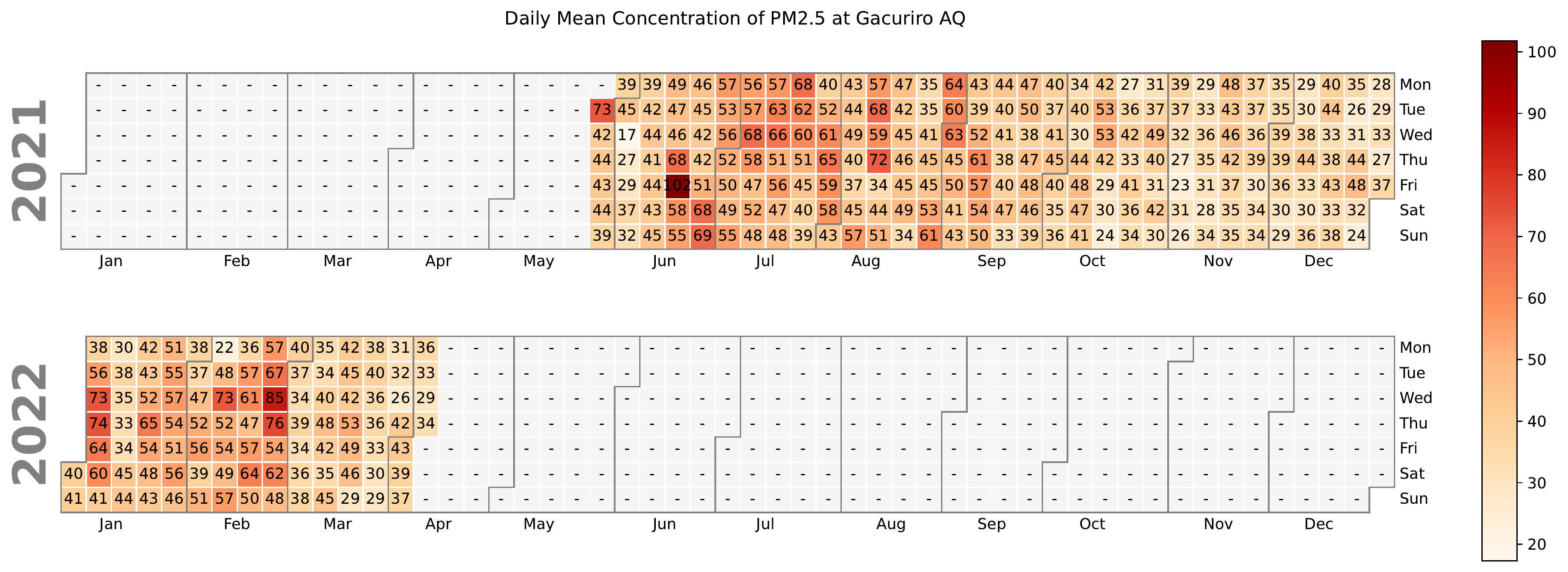}  
		\label{fig:sub-second_calendar}
\end{figure}
\FloatBarrier
\begin{figure}[ht!]
    \centering
		\caption{Calendar Plots of Daily PM2.5 Concentration  at Rebero}
		\includegraphics[width=1.02\linewidth]{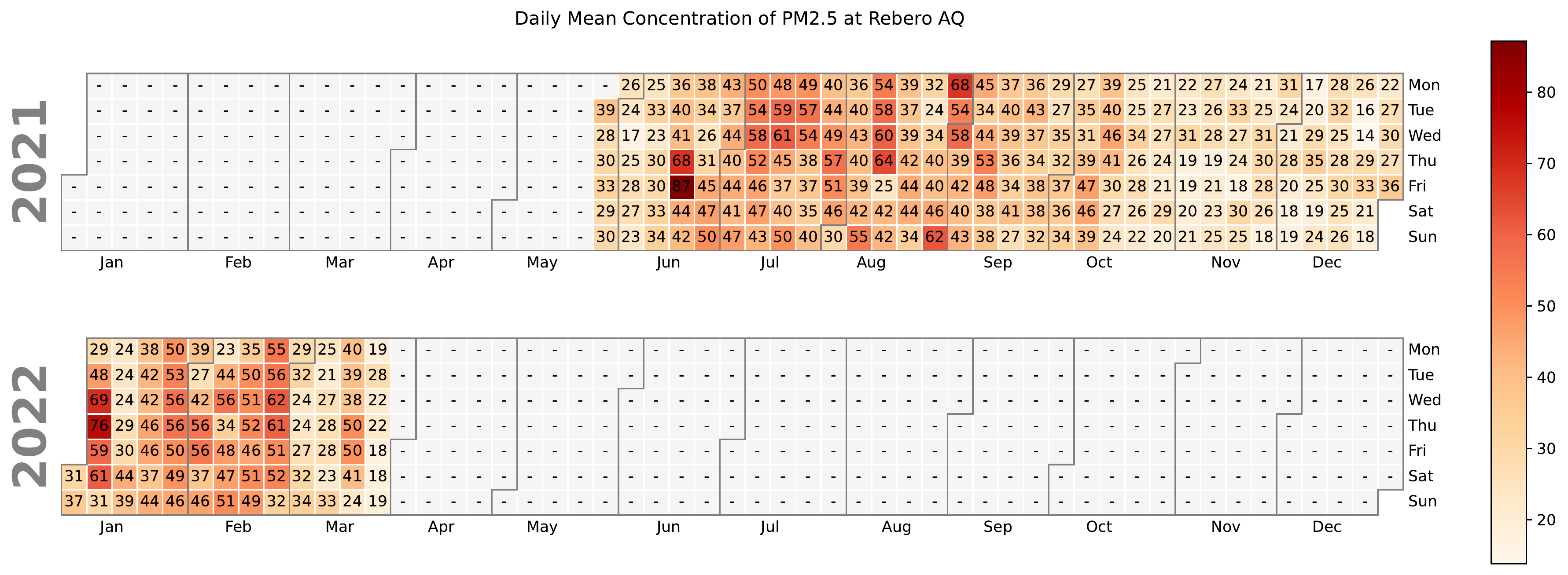}  
		\label{fig:sub-third_calendar}
\end{figure}
\FloatBarrier
\begin{figure}[ht!]
    \centering
		\caption{Calendar Plots of Daily PM2.5 Concentration  at Kimihurura}
		\includegraphics[width=1.02\linewidth]{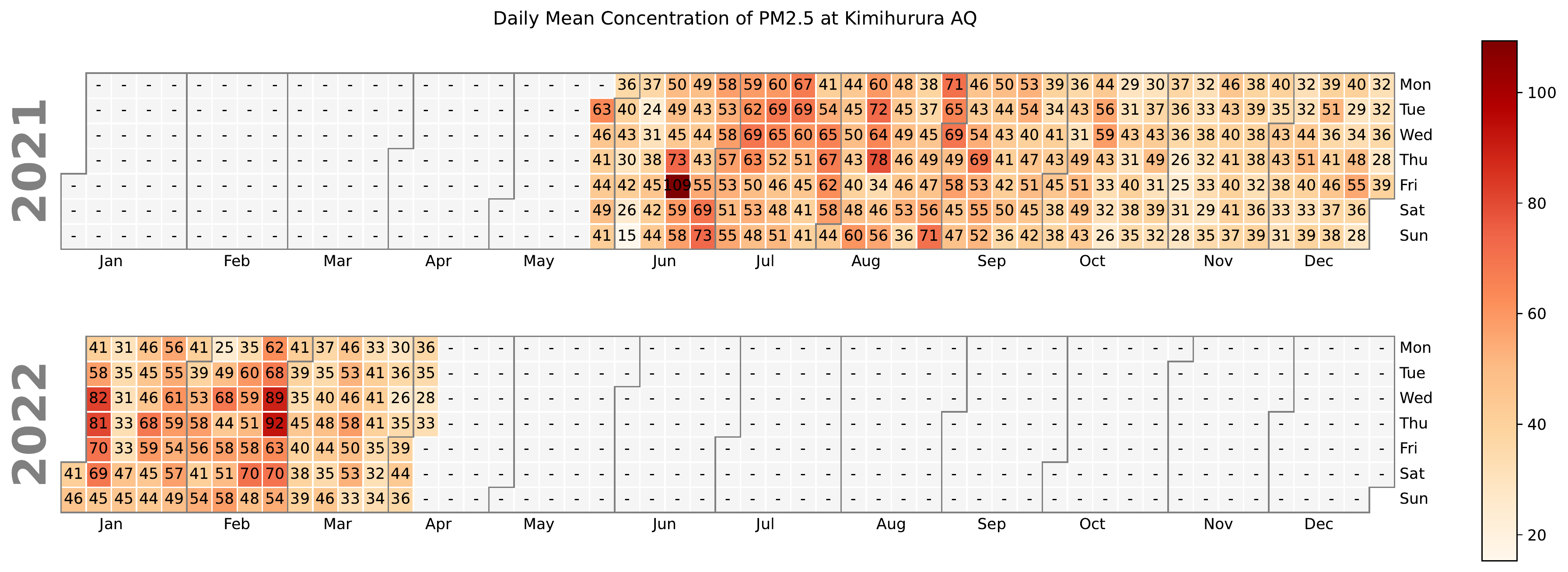}  
		\label{fig:sub-fourth_calendar}
\end{figure}
\FloatBarrier
\begin{figure}[ht!]
    \centering
		\caption{Calendar Plots of Daily PM2.5 Concentration  at Gikondo}
		\includegraphics[width=1.02\linewidth]{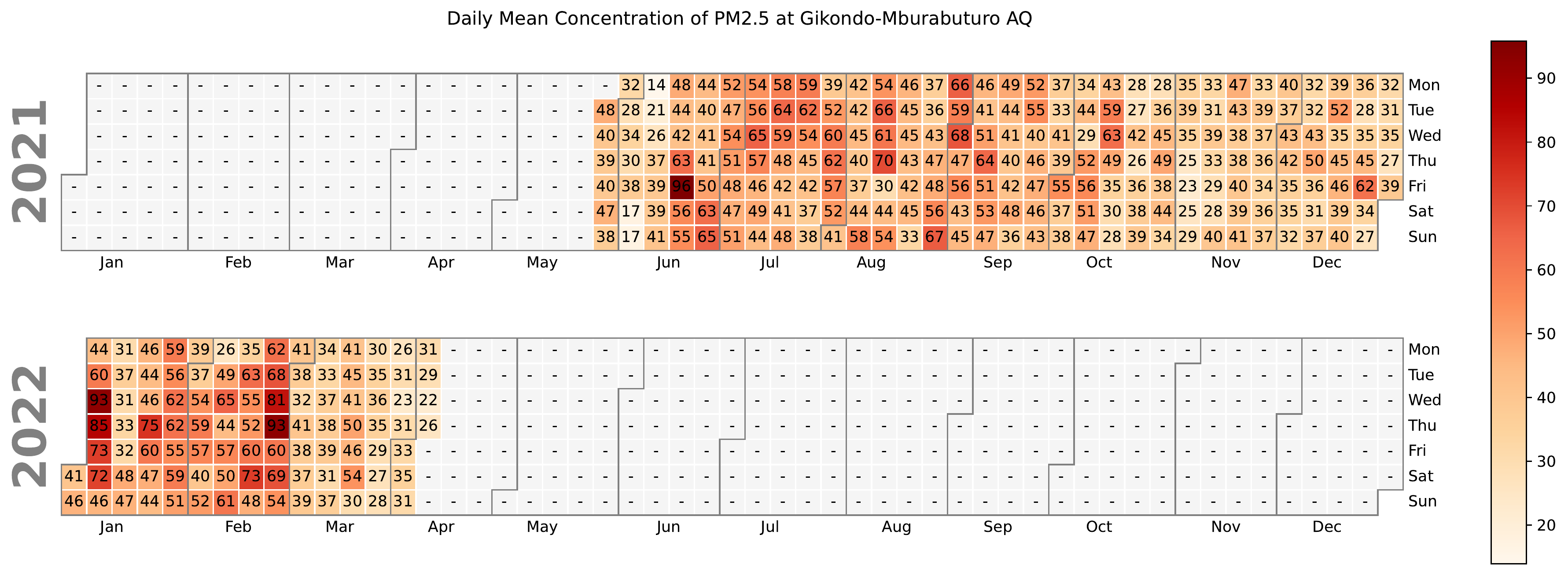}  
		\label{fig:sub-fifth_calendar}
\end{figure}
\FloatBarrier
\begin{figure}[ht!]
    \centering
		\caption{Calendar Plots of Daily PM2.5 Concentration  at Kiyovu}
		\includegraphics[width=1.02\linewidth]{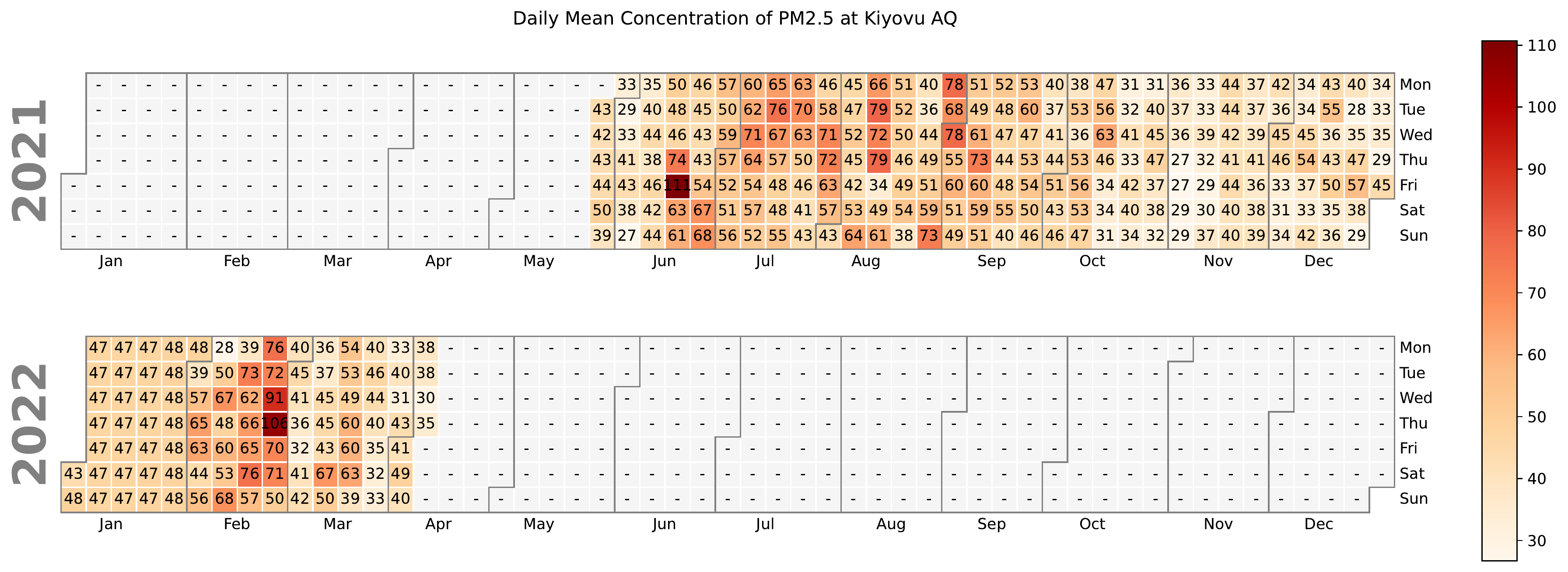}  
		\label{fig:sub-sixth_calendar}
\end{figure}
\FloatBarrier
\begin{figure}[ht!]
    \centering
		\caption{Calendar Plots of Daily PM2.5 Concentration  at Mount Kigali}
		\includegraphics[width=1.02\linewidth]{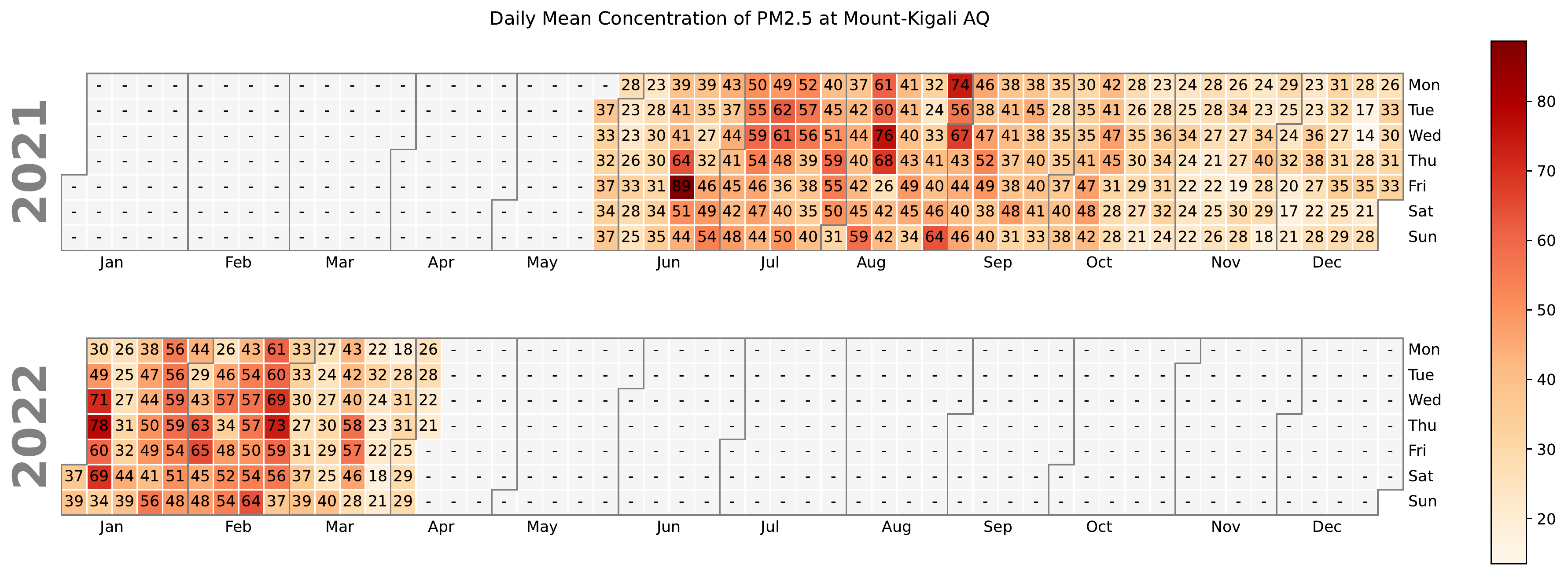}
		\label{fig:sub-seventh_calendar}
\end{figure}
\FloatBarrier
\begin{figure}[ht!]
    \centering
		\caption{Calendar Plots of Daily PM2.5 Concentration  at Rusororo}
		\includegraphics[width=1.02\linewidth]{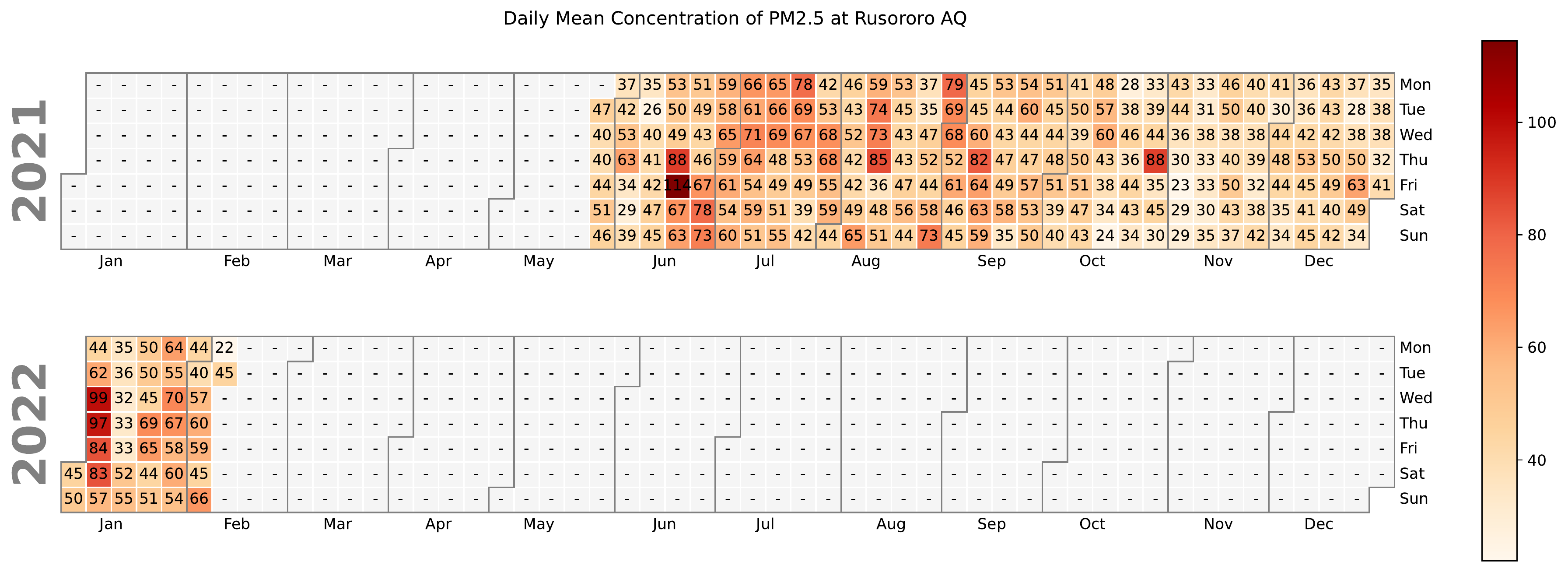}  
		\label{fig:sub-eighth_calendar}
\end{figure}
\FloatBarrier
\begin{figure}[ht!]
    \centering
	\caption{Calendar Plots of Daily PM2.5 Concentration  at Gikomero}
	\includegraphics[width=1.05\linewidth]{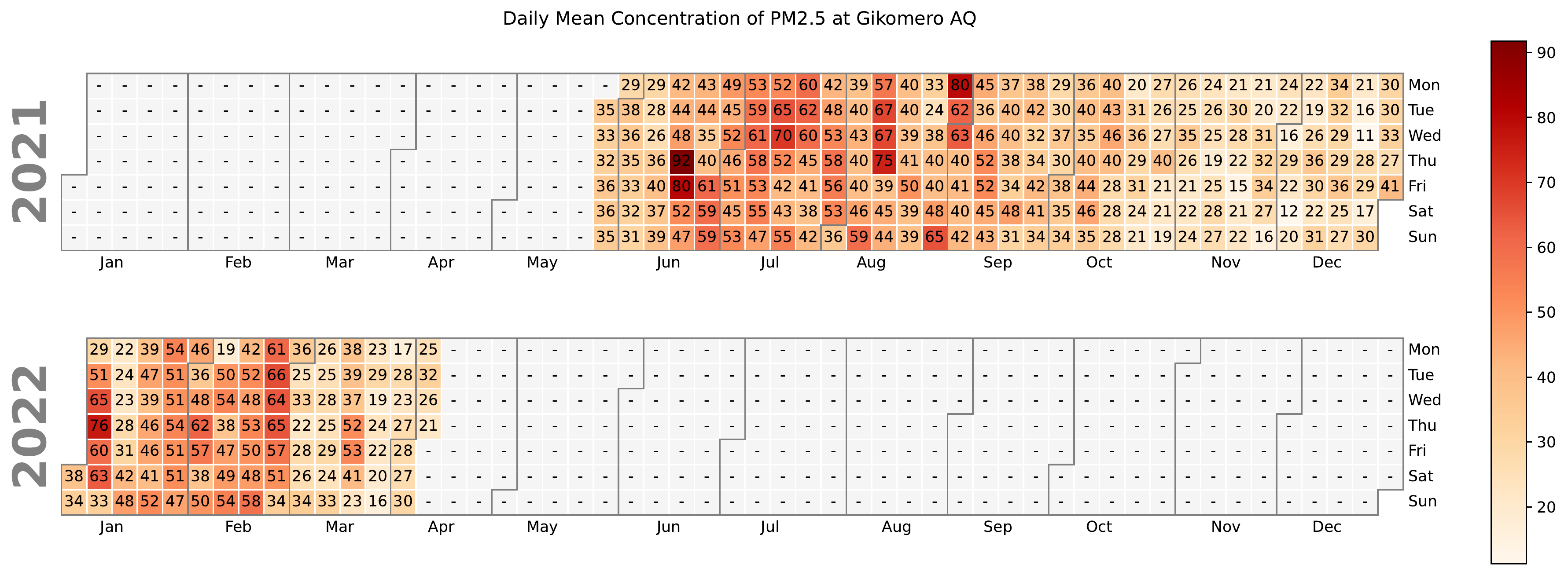}  
	\label{fig:sub-ninth_calendar}
\end{figure}
\FloatBarrier
The seasonal trend analysis is connected to the climatic seasons found in Rwanda. Rwanda, a sub-Saharan country with a tropical climate, has a hilly landscape, particularly in the east and west. The climate of the country is divided into four seasons: the long dry season (June to August), the short rainy season (September to November), the short dry season (December to February), and the long rainy season (March to May). The amount of air pollutants, particularly PM2.5 emitted varies according to the activities carried out during each season. During the long dry season, the highest levels of PM2.5 were found in June to July $2021$. Agricultural and construction activities are other major sources that contribute to the reduction of the air quality through dust. During these months, a large amount of dust, including larger or coarser particles known as PM10 (between $2.5$ and $10$ micrometres) are released in the atmosphere . PM10, like PM2.5, is hazardous as it causes nasal and throat irritation as well as allergic reactions when inhaled while breathing. The average PM2.5 concentration calculated at all stations from June to August $2021$ (\ref{fig:sub-first_Seasonal}) was $\SI{45.844}{\micro\gram}/m^3$, whereas in other seasons, average seasonal $PM2.5$ concentrations of $\SI{37.358}{\micro\gram}/m^3$, $\SI{44.155}{\micro\gram}/m^3$ and $\SI{35.063}{\micro\gram}/m^3$ were found in the short rainy season, short dry season, and long rainy season respectively.

\begin{figure}[ht!]
    \centering
		\caption{Short dry season (December to February)}
		\includegraphics[width=1.01\linewidth]{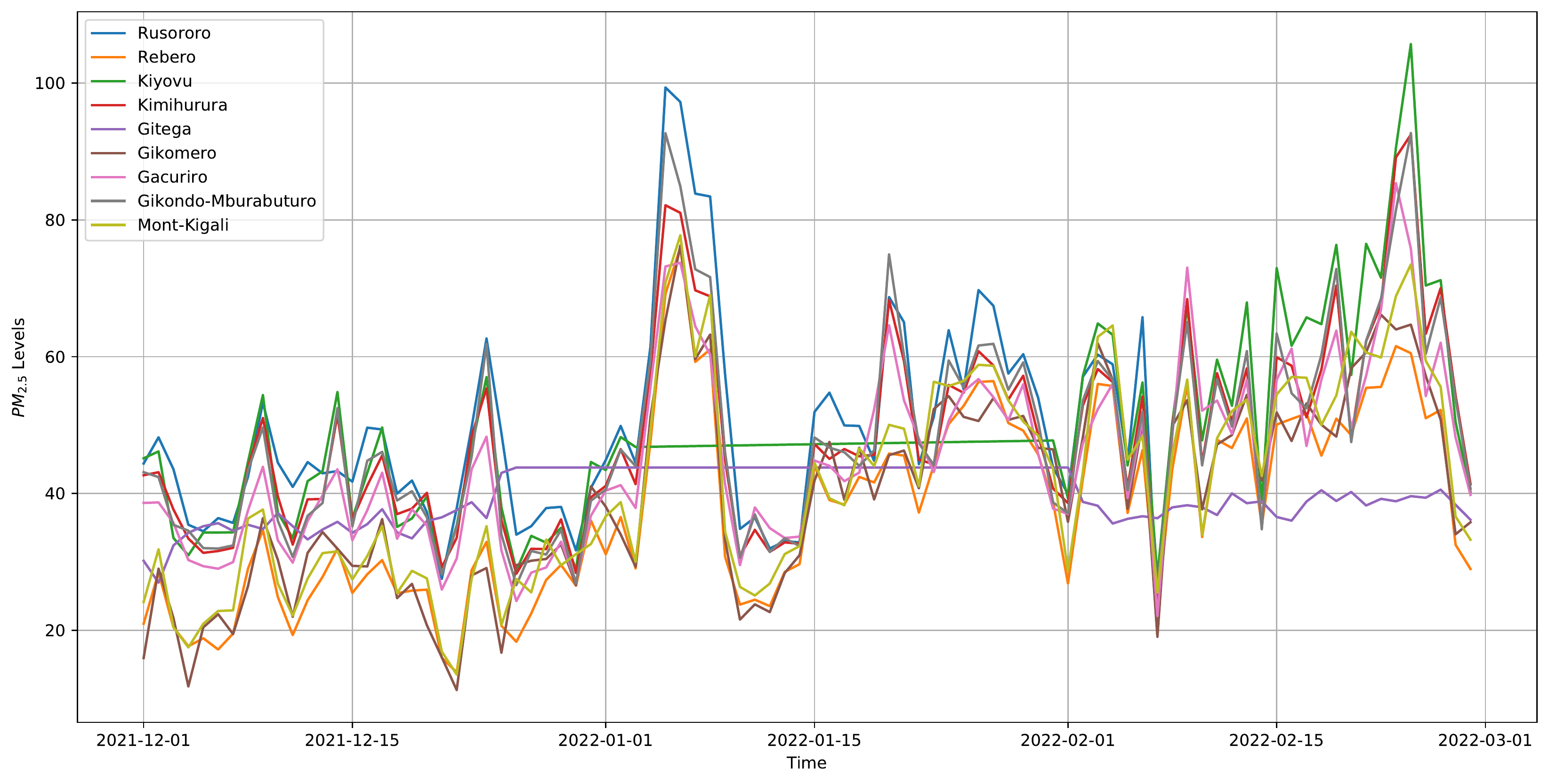}  
		\label{fig:sub-third_Seasonal}
\end{figure}

In the short dry season (\ref{fig:sub-third_Seasonal}), the highest mean concentration of PM2.5 was found in January $2022$ at all stations as a result  of different factors: high school students are returning to school after holidays, and many workers in this month are returning to work after  festive season, which contribute to the increase in pollution levels.
\FloatBarrier
\begin{figure}[ht!]
    \centering
		\caption{Short rainy season (September to November)}
		\includegraphics[width=1.01\linewidth]{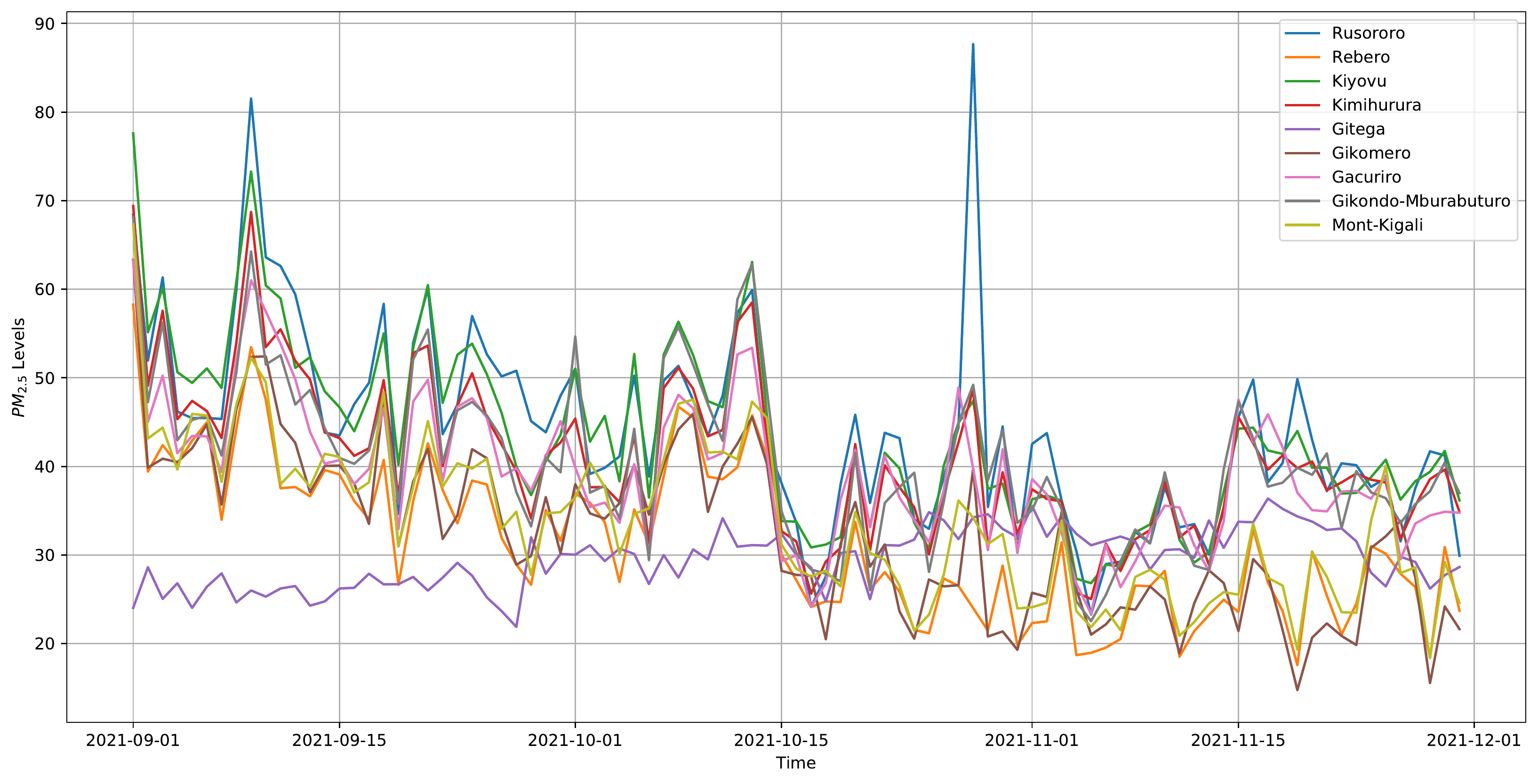}  
		\label{fig:sub-second_Seasonal}
\end{figure}
\FloatBarrier
\FloatBarrier
\begin{figure}[ht!]
    \centering
		\caption{Long rainy season (March to May)}
		\includegraphics[width=1.01\linewidth]{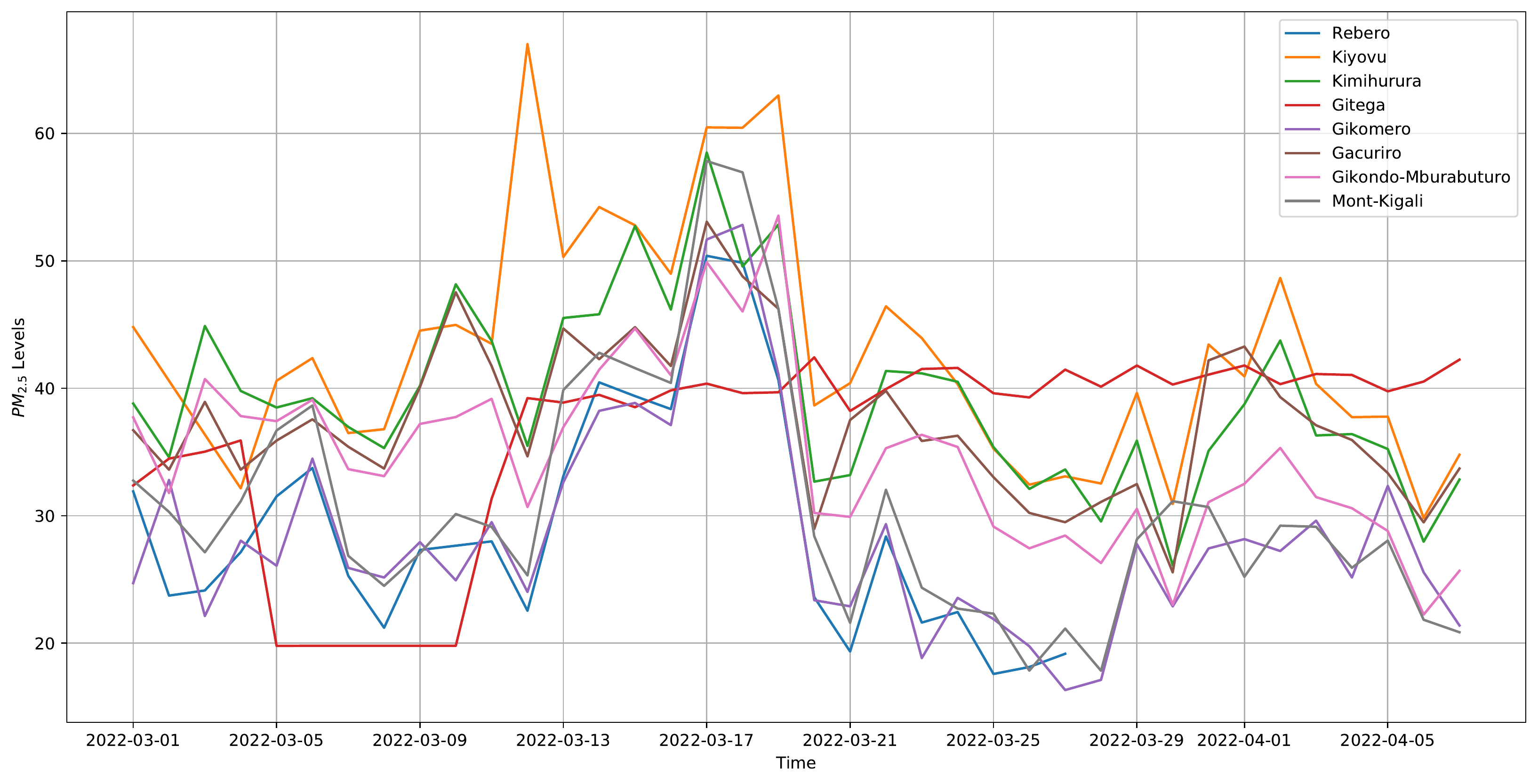}  
		\label{fig:sub-fourth_Seasonal}
\end{figure}
\FloatBarrier
	

\subsection{Forecasting}

\begin{figure}[ht!]
    \centering
  	\caption{Forecasting daily PM2.5 concentrations at Kiyovu}
		\includegraphics[width=1.02\linewidth]{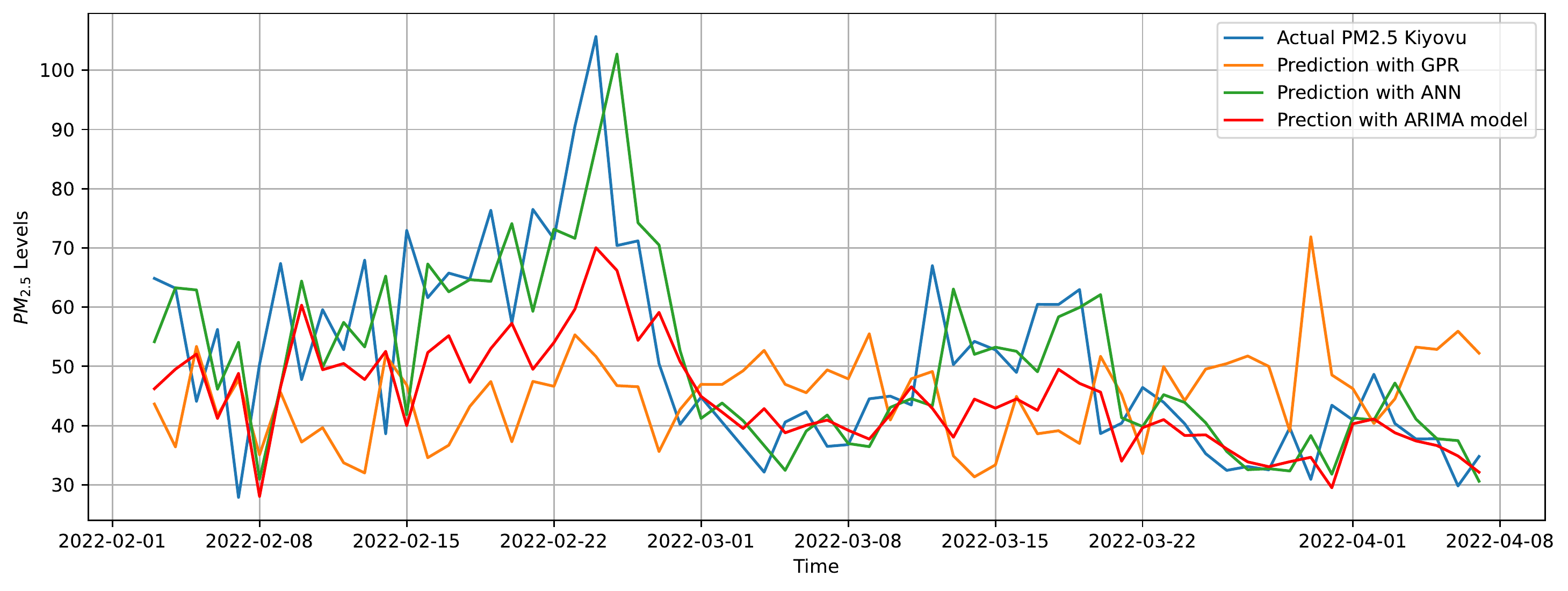}  
		\label{fig:sub-sixth_Forecasting}
\end{figure}

\begin{figure}[ht!]
    \centering
		\caption{Forecasting daily PM2.5 concentrations at \\Gikomero}
		\includegraphics[width=1.02\linewidth]{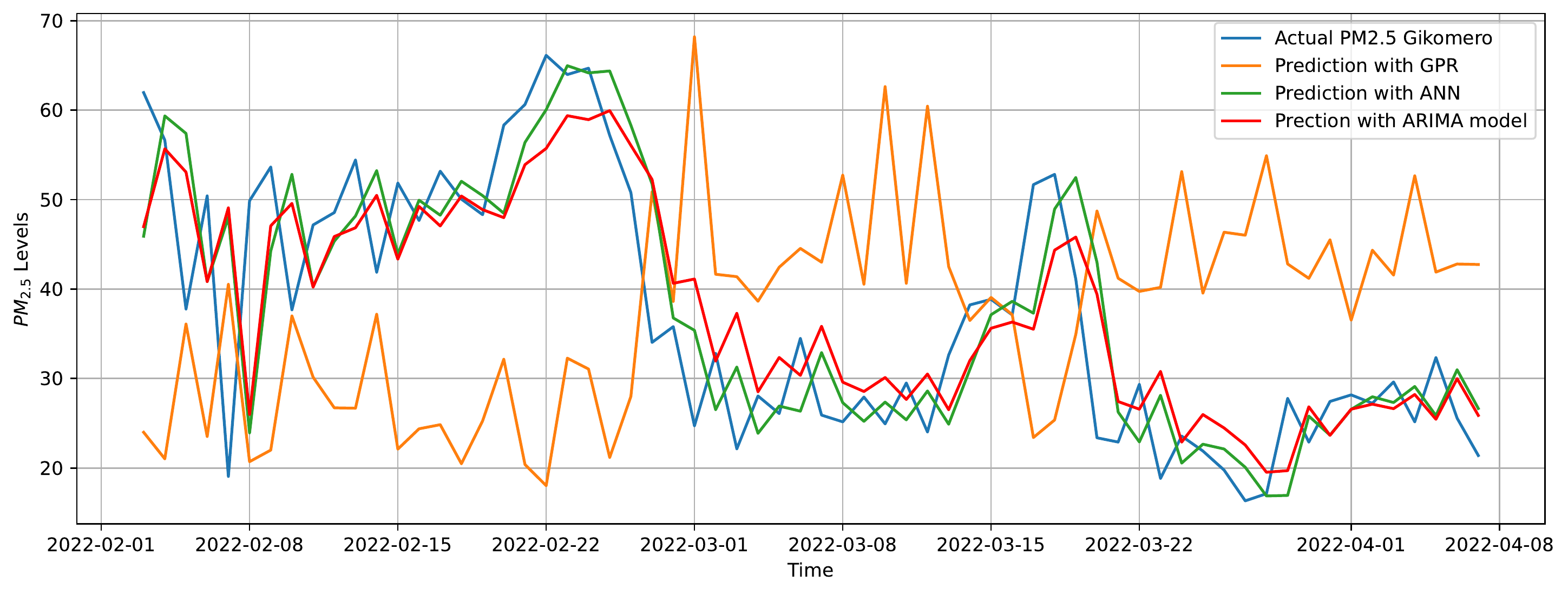}
		\label{fig:sub-ninth_Forecasting}
\end{figure}

The forecasting plots against daily PM2.5 concentrations for the Autoregressive Integrated Moving Average, Artificial Neural Network, and Gaussian Process Regression models are shown in figures (\ref{fig:sub-sixth_Forecasting},$\cdots$,\ref{fig:sub-ninth_Forecasting}). All the PM2.5 concentrations measured at each of the air quality monitoring stations were modeled using all three models. The models were compared using statistical metrics: Root Mean square Error (RMSE) and Mean Absolute Error (MAE). 
\begin{align*}
	RMSE= \sqrt{\frac{1}{N}\sum_{i=1}^N(y_i-\hat{y_i})^2}
\end{align*}
\begin{align*}
	MAE= \frac{1}{N}\sum_{i=1}^N |y_i-\hat{y_i}|
\end{align*}
where $N$ denotes the total number of observations, $y_i$  denote the observed values and $\hat{y_i}$ are model predictions.
\FloatBarrier
\begin{figure}[ht!]
    \centering
\caption{Forecasting daily PM2.5 concentrations at Gacuriro}
		\includegraphics[width=1.02\linewidth]{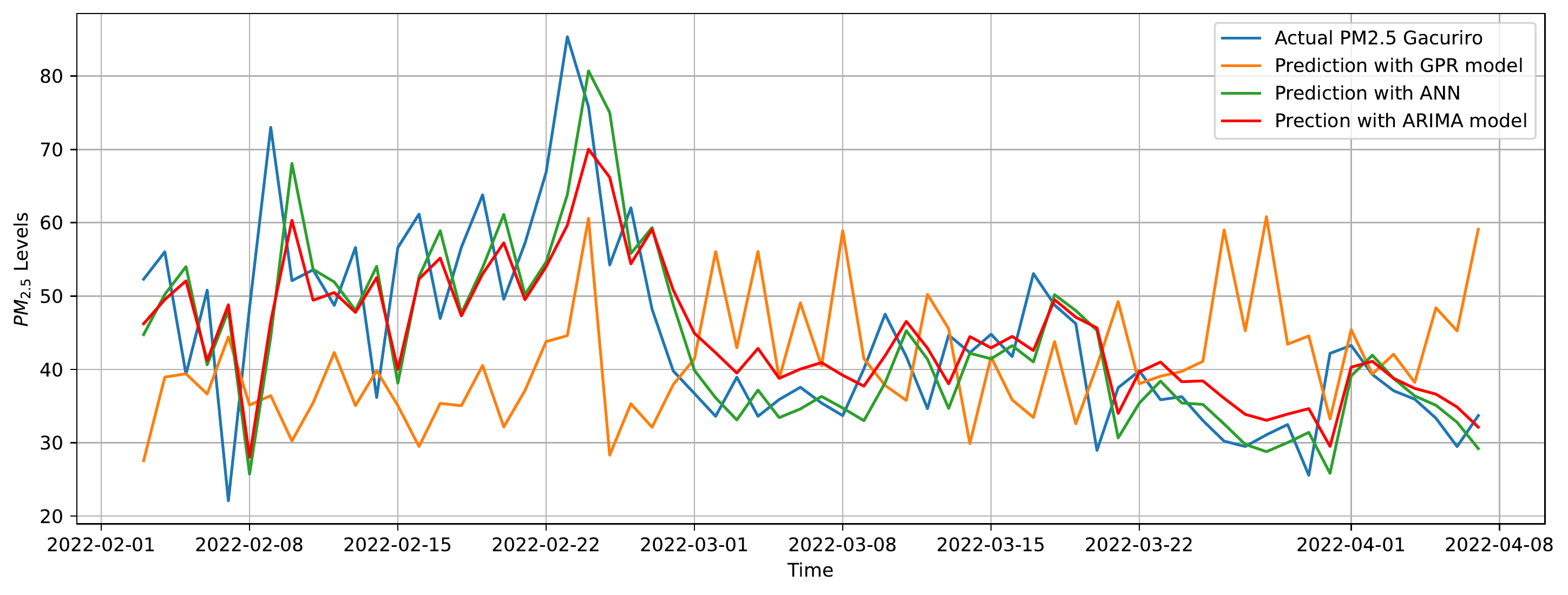} 
		\label{fig:sub-second_Forecasting}
\end{figure}
\FloatBarrier
\begin{figure}[ht!]
    \centering
\caption{Forecasting daily PM2.5 concentrations at Rebero}
		\includegraphics[width=1.02\linewidth]{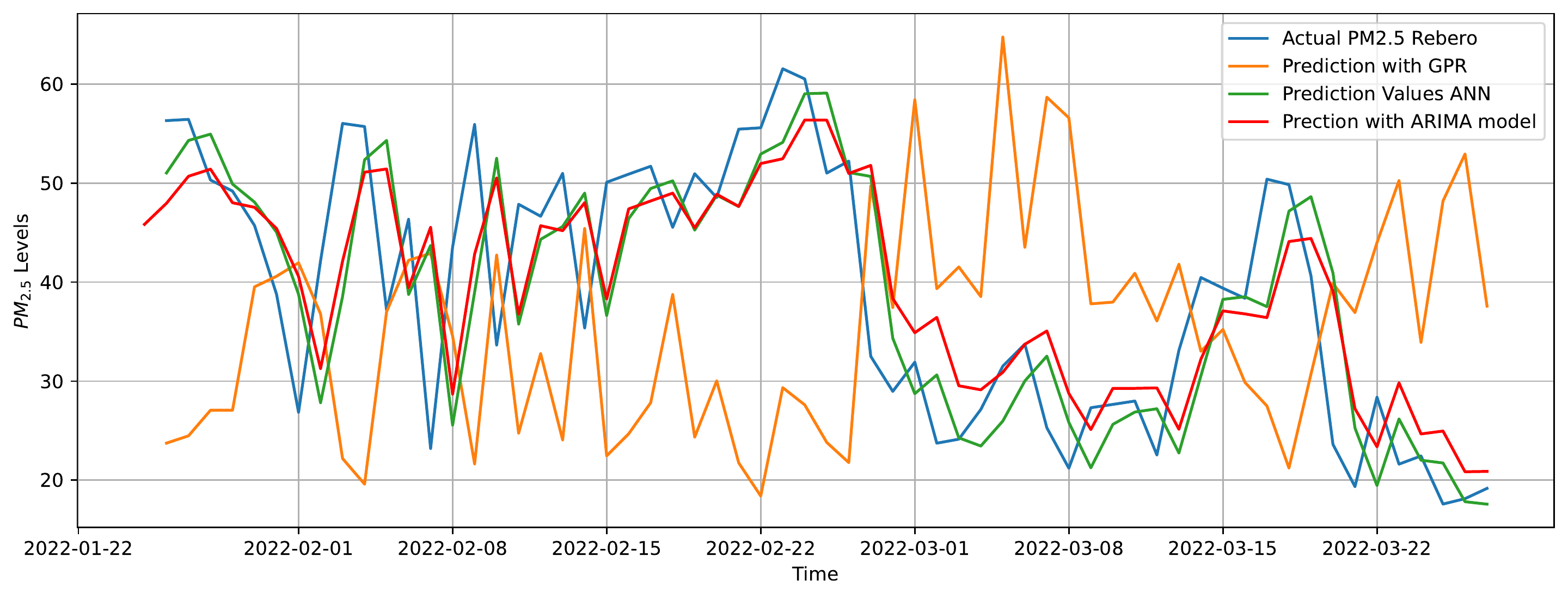}  
		\label{fig:sub-third_Forecasting}
\end{figure}
\FloatBarrier
\begin{figure}[ht!]
    \centering
		\caption{Forecasting daily PM2.5 concentrations at Kimihurura}
		\includegraphics[width=1.02\linewidth]{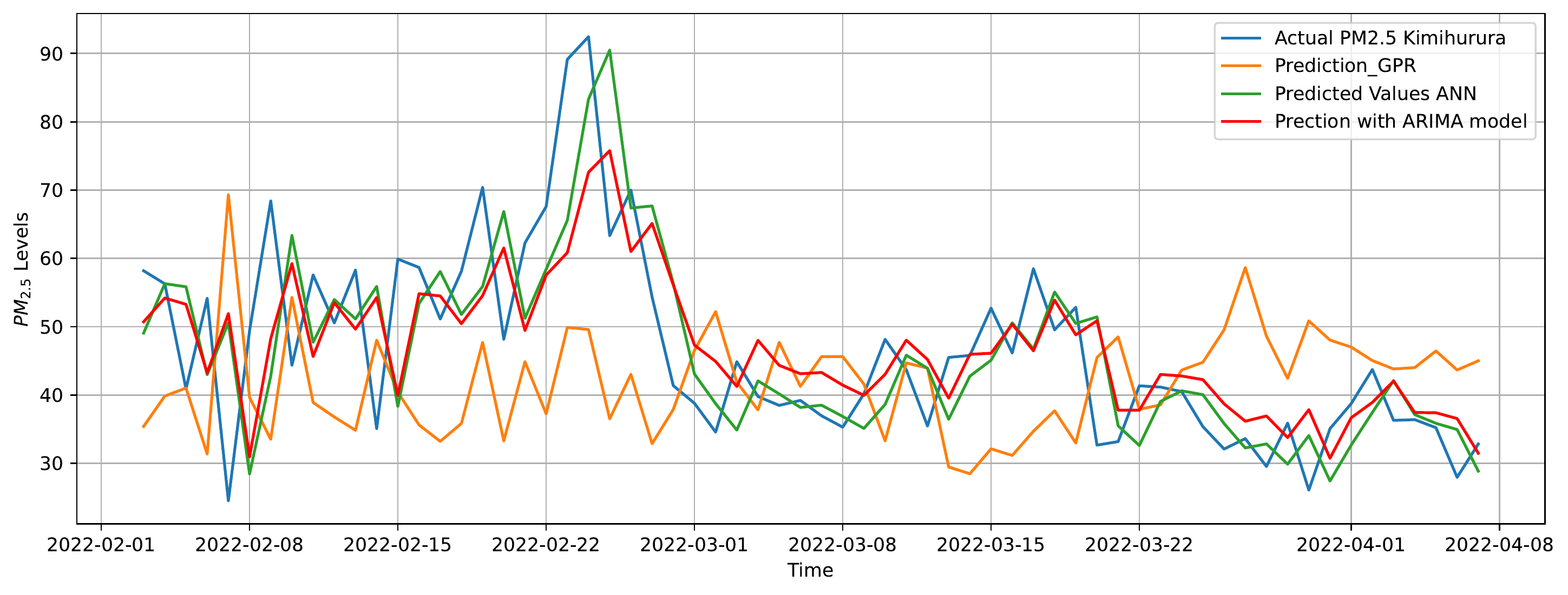}  
		\label{fig:sub-fourth_Forecasting}
\end{figure}
\FloatBarrier
\begin{figure}[ht!]
    \centering
		\caption{Forecasting daily PM2.5 concentrations at Gikondo-Mburabuturo}
		\includegraphics[width=1.02\linewidth]{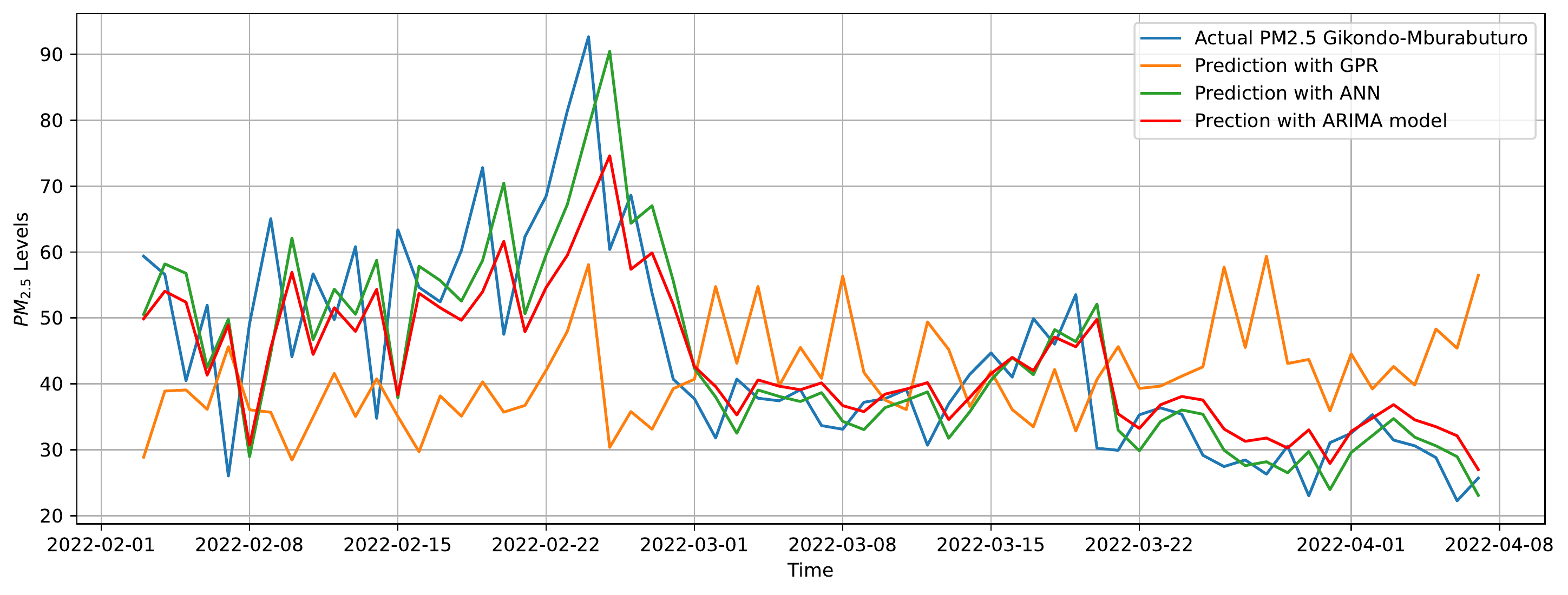}  
		\label{fig:sub-fifth_Forecasting}
\end{figure}
\FloatBarrier

\begin{figure}[ht!]
    \centering
		\caption{Forecasting daily PM2.5 concentrations at Rusororo}
		\includegraphics[width=1.02\linewidth]{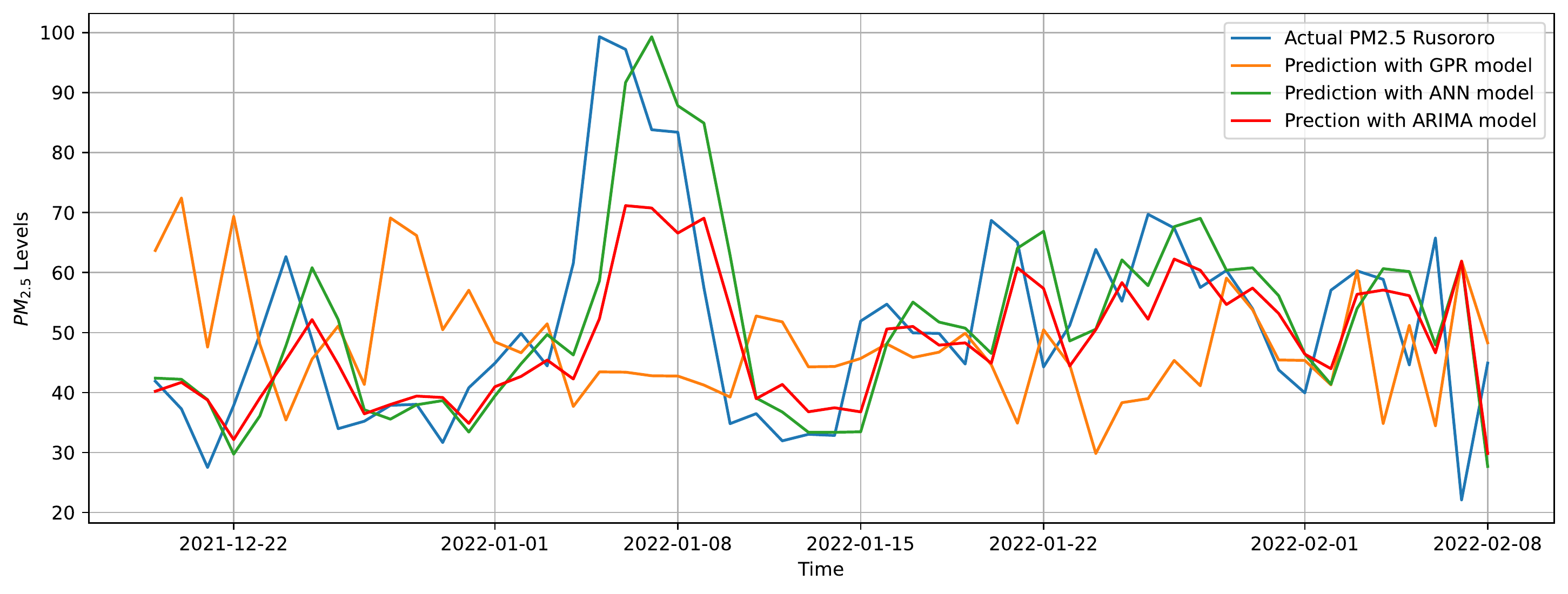}	\label{fig:sub-eighth_Forecasting}
\end{figure}
\FloatBarrier
\begin{figure}[ht!]
    \centering
		\caption{Forecasting daily PM2.5 concentrations at Mount Kigali}
		\includegraphics[width=1.02\linewidth]{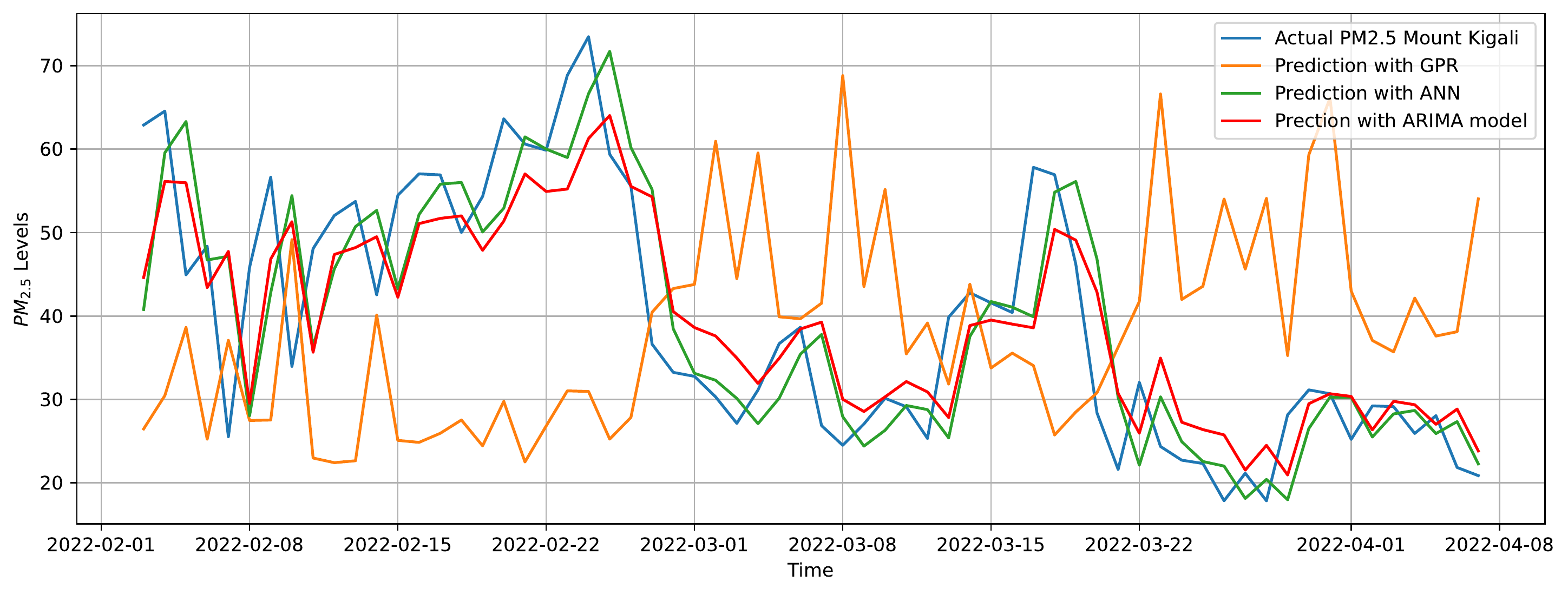}  
		\label{fig:sub-seventh_Forecasting}
\end{figure}
\FloatBarrier
\begin{figure}[ht!]
    \centering
		\caption{Forecasting daily PM2.5 concentrations at \\Gitega}
		\includegraphics[width=1.02\linewidth]{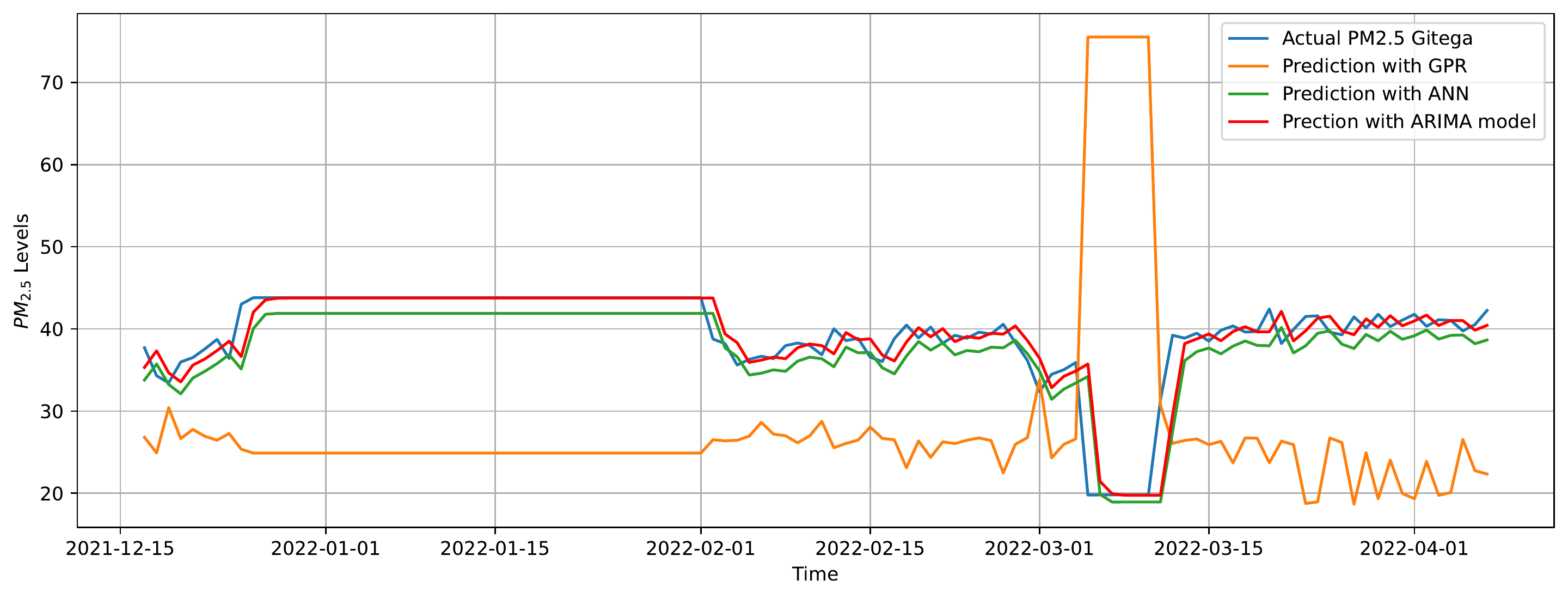}  
		\label{fig:sub-first_Forecasting}
\end{figure}
\FloatBarrier
\FloatBarrier
Table (\ref{tab:my_label}) shows the RMSE and MAE calculated while predicting PM2.5 concentrations measured at each air quality monitoring station installed in Kigali city. The RMSEs found using the ARIMA model were the lowest values compared to the RMSEs found using the ANN and GPR models, as shown in the table(\ref{tab:my_label}). This demonstrates that the ARIMA model suited the PM2.5 time series data the best. The ANN had the lowest RMSE of any other model in the PM2.5 time series at Gikomero air monitoring station. In terms of MAE, the ARIMA model produced the lowest  MAE values when compared to other models. 
\begin{table}[ht!]
\caption{The RMSEs and MAEs of ARIMA, ANN and GPR models.}
	\centering
	\resizebox{\columnwidth}{!}{%
	\begin{tabular}{ ||p{1.82cm}||p{0.8cm}|p{0.8cm}|p{0.8cm}||p{0.8cm}|p{0.8cm}|p{0.8cm}||  }
		\hline
		\hline
		\multicolumn{1}{||c||}{\textbf{Stations}} &\multicolumn{3}{c||}{\textbf{RMSE}} &
		\multicolumn{3}{c||}{\textbf{MAE}}\\
		\hline
		& \textbf{{\footnotesize ARIMA}} &\textbf{{\footnotesize ANN}}&\textbf{{\footnotesize GPR}}& \textbf{{\footnotesize ARIMA}} &\textbf{{\footnotesize ANN}}&\textbf{{\footnotesize GPR}}\\
		\hline
		Gacuriro AQ   & 8.942    & 10.183 &   17.184& 7.489    &7.700&   16.886\\
		\hline
		Gikomero AQ&   8.942  & 8.926   &24.215& 6.800    &6.681&   21.200\\
		\hline
		Gikondo Mburabuturo AQ &10.512 & 10.842&  16.694&  7.967    &8.092&   16.886\\
		\hline
		Gitega AQ   &2.537 & 2.746&  20.162& 1.241    &1.694&   11.960\\
		\hline
		Kimihurura AQ&   10.621  & 11.071&18.353& 8.532    &8.542&   14.979\\
		\hline
		Kiyovu AQ& 11.875 & 12.403   &20.162& 9.312    &9.192&   17.195\\
		\hline
		Mount Kigali AQ& 8.701  & 9.361&24.437& 6.905    &6.942&   21.636\\
		\hline
		Rebero AQ& 8.606  & 8.888&22.286& 6.915    &6.774&   18.158\\
		\hline
		Rusororo AQ& 13.464  & 13.747&22.409& 9.840    &10.149&   17.474\\
		\hline
		\hline
	\end{tabular}
}
	\label{tab:my_label}
\end{table}
\FloatBarrier
\section{Conclusion}

This study has demonstrated that Autoregressive Integrated Moving Average, Artificial Neural Network, and Gaussian Process Regression models can be used to forecast PM2.5 data in Rwanda. Future works should consider other air pollutants such as PM10, SO2, NOx, CO, and O3. This would assist policymakers and agencies in charge of environmental and health protection in identifying best mitigation strategies to air pollution-related problems and preventing citizens from being exposed to unhealthy air.









 \section{Acknowledgments}
 This work was funded by the African Institute for Mathematical Sciences with the financial support from the government of Canada provided through Global Affairs Canada and the International Development Research Centre.

\bibliographystyle{abbrv}
\bibliography{references}




\end{document}